\documentclass[10pt,twocolumn,letterpaper]{article}

\usepackage[pagenumbers]{iccv} 

\newcommand{\todo}[1]{{\color{red}#1}}

\usepackage{graphicx}
\usepackage{mathtools}

\definecolor{iccvblue}{rgb}{0.21,0.49,0.74}
\usepackage[pagebackref,breaklinks,colorlinks,allcolors=iccvblue]{hyperref}

\usepackage[accsupp]{axessibility}  

\def\confYear{2025}

\title{Monocular Facial Appearance Capture in the Wild}

\author{Yingyan Xu$^{1, 2}$
	\hspace{7mm}
	Kate Gadola$^{1}$
	\hspace{7mm}
	Prashanth Chandran$^{2}$
	\\
	Sebastian Weiss$^{2}$
	\hspace{7mm}
	Markus Gross$^{1, 2}$
	\hspace{7mm}
	Gaspard Zoss$^{2}$
	\hspace{7mm}
	Derek Bradley$^{2}$
	\\
	$^{1}$ETH Z\"urich
	\hspace{7mm}
	$^{2}$DisneyResearch\textbar Studios \\
	{\tt\small {\{yingyan.xu,grossm\}@inf.ethz.ch}} 
	\hspace{7mm} {\tt\small kgadola@student.ethz.ch} \\ 
	{\tt\small {\{prashanth.chandran,sebastian.weiss,gaspard.zoss,derek.bradley\}@disneyresearch.com}}
}

\begin{document}
\maketitle

\newcommand{\figref}[1]{Fig.~\ref{#1}}
\newcommand{\tabref}[1]{Table~\ref{#1}}
\newcommand{\eqnref}[1]{Eq.~\ref{#1}}
\newcommand{\secref}[1]{Section~\ref{#1}}
\newcommand{\appref}[1]{Appendix~\ref{#1}}


\newcommand{\shortcite}[1]{\cite{#1}}


\definecolor{dbcolor}{RGB}{50,10,210}
\newcommand\db[1] {{\textcolor{dbcolor}{\em\textbf{DB}: #1}}}

\definecolor{piccolo}{RGB}{10,150,100}
\newcommand\pc[1] {{\textcolor{pccolor}{\em\textbf{PC}: #1}}}

\definecolor{swcolor}{RGB}{210,10,210}
\newcommand\sw[1] {{\textcolor{swcolor}{\em\textbf{SW}: #1}}}

\definecolor{gzcolor}{RGB}{210,210,10}
\newcommand\gz[1] {{\textcolor{gzcolor}{\em\textbf{GZ}: #1}}}

\definecolor{yxcolor}{RGB}{10,210,50}
\newcommand\yx[1] {{\textcolor{yxcolor}{\em\textbf{YX}: #1}}}

\definecolor{delcolor}{RGB}{210,0,0}
\definecolor{addcolor}{RGB}{0,0,0}
\newcommand\del[1] {{\textcolor{delcolor}{#1}}}
\newcommand\add[1] {{\textcolor{addcolor}{#1}}}

\renewcommand\todo[1] {{\textcolor{red}{\em\textbf{TODO}: #1}}}
\newcommand\camready[1] {{\textcolor{blue}{#1}}}

\makeatletter
\newcommand\footnoteref[1]{\protected@xdef\@thefnmark{\ref{#1}}\@footnotemark}
\makeatother

\renewcommand{\thefootnote}{\fnsymbol{footnote}}

\clubpenalty=10000
\widowpenalty=10000
\displaywidowpenalty=10000

\begin{abstract}

We present a new method for reconstructing the appearance properties of human faces from a lightweight capture procedure in an unconstrained environment.  Our method recovers the surface geometry, diffuse albedo, specular intensity and specular roughness from a monocular video containing a simple head rotation in-the-wild.  Notably, we make no simplifying assumptions on the environment lighting, and we explicitly take visibility and occlusions into account.  As a result, our method can produce facial appearance maps that approach the fidelity of studio-based multi-view captures, but with a far easier and cheaper procedure.

\end{abstract}    
\section{Introduction}
\label{sec:intro}

3D facial scanning is a fundamental tool for the creation of realistic digital humans in several industries like film and video game entertainment, communication and telepresence, medical applications, and the new trend of AI-driven digital characters.  For decades, practitioners have relied on high-quality 3D face scans in order to bring people into virtual worlds.  Much of the technology evolution has focused on reconstructing the surface geometry, where initial scanners could create detailed triangle meshes in controlled studio settings with many cameras and lights, and then more recent efforts focused on fast and lightweight face reconstruction from monocular inputs in less-constrained, so-called ``in-the-wild" settings.  While the facial geometry is extremely important, the shape alone is not enough to re-render the subject in novel environments with photorealistic quality.  For this task, we must additionally recover the appearance properties of the face, which dictate how light interacts with the skin surface.  As such, in today's high-end facial scanning pipelines the desired result includes a high-resolution facial surface mesh with corresponding appearance textures for properties like the diffuse albedo, specular intensity and specular roughness, which are compatible with modern skin shaders.  

Like facial geometry reconstruction, the field of skin appearance estimation is also well-studied in controlled studio environments, where accurate appearance maps can be reconstructed from large setups that obtain multi-view or multi-shot images under calibrated lighting~\cite{riviere2020single}.  Following the geometry trend, current research aims to allow facial appearance capture in less constrained settings, for example outdoors using the sun as a single point light~\cite{wang2023sunstage}.  Unfortunately, these methods often make simplifying assumptions, and thus there still exists a large gap in reconstruction quality between current in-the-wild methods and production-ready studio-based capture.  

In this work we present a new method for facial appearance capture in the wild, surpassing the level of fidelity of existing lightweight methods.  Our approach requires only a short video sequence of a simple head rotation, captured from a single camera in any environment, including indoors or outdoors, on a sunny day or in shadow.  Our approach is built on traditional inverse rendering optimization, where a fast differentiable renderer is used to solve for the geometry and appearance parameters together with the environment lighting simultaneously.  Different from previous methods, we do not make any assumptions on the lighting condition (\eg we do not require a sun in the sky), and as our main contribution we explicitly take visibility into account, effectively removing baked-in shading by correctly modeling self-occlusion in our appearance solver. The result is a detailed geometry mesh with textures for diffuse albedo, specular intensity and roughness. 
As we will show, our approach leads to more faithful recovery of the appearance properties than existing techniques in the wild.  As a particular application, our approach allows fast capture of actors on a film set, with resulting assets that can be used directly in traditional VFX pipelines. In summary, we make the following key contributions:
\begin{itemize}
	\item A new state-of-the-art method for in-the-wild facial appearance capture that makes no assumption on the scene lighting condition.
	\item A novel shading model that explicitly handles visibility and self-occlusion for inverse rendering pipelines, achieving high-quality appearance reconstruction from monocular input. 
\end{itemize}
\section{Related Work}
\label{sec:related-work}

In the following section, we first outline relevant works around in-the-wild inverse rendering which do not necessarily focus on the human face. Second, we highlight works which specifically tackle facial appearance capture.

\vspace{-3mm}
\paragraph{Inverse rendering in the wild.}

Inverse rendering is the process of decomposing the scene into 3D shape, material and illumination by simulating the rendering process and comparing the results against captured images. It has been a popular research topic with the recent advances in novel view synthesis using neural implicit representations~\cite{mildenhall2021nerf, muller2022instant, wang2021neus}, 3D Gaussian splatting~\cite{kerbl3Dgaussians}, mesh-based differentiable renderers (rasterizers~\cite{Laine2020diffrast} and path tracers~\cite{Li:2018:DMC, Mitsuba3}) and using diffusion models as prior~\cite{lyu2023dpi, Yenyo2024, liang2024photorealistic, chen2024intrinsicanything}. While some existing techniques target a very challenging scenario where the lighting can differ across different images~\cite{boss2021neural, boss2022samurai, engelhardt2024shinobi}, we restrict our discussion here to methods that assume a static unknown environment lighting. PhySG~\cite{physg2021} utilizes spherical Gaussians to approximately and efficiently evaluate the rendering equation in closed form. Munkberg \etal~\cite{Munkberg_2022_CVPR} propose an efficient end-to-end framework for joint learning of topology, triangle meshes and materials, achieving much faster training and inference compared to previous NeRF-based factorization methods~\cite{zhang2021nerfactor, boss2021nerd, srinivasan2021nerv}. They also introduce a differentiable formulation of the split-sum approximation of environment lighting to efficiently recover all-frequency lighting. Follow-up work~\cite{hasselgren2022nvdiffrecmc} shows that material and lighting decomposition can be further improved with a more realistic shading model, incorporating ray tracing and Monte Carlo integration.
Recently, 3D Gaussian Splatting techniques were used in conjunction with physical-based rendering to allow for scene relighting~\cite{liang2024gs, gao2023relightable, bi2024rgs,  zhu2024gs, wu2024deferredgs}.

\vspace{-3mm}
\paragraph{Facial appearance capture.}
Traditional face capture studios often employ a multi-view setup with controlled and calibrated lighting conditions to reconstruct the skin appearance~\cite{riviere2020single, ghosh2011multiview, gotardo2018practical, debevec2000acquiring}. Similar setups were used for facial appearance decomposition with neural and gaussian primitives~\cite{saito2024relightable, yang2023towards, Xu_2024_CVPR, Xu_2023_ICCV, sarkar2023litnerf}. Lighter alternative setups were also explored by Lattas \etal~\cite{lattas2022practical} and Choi \etal~\cite{choi2024differentiable} but they still require the subject to be seating in a dedicated space. Recently, the research community has investigated more lightweight setups which are easily accessible to everyone. However, most of these techniques still pose some constraints on the capture environment. CoRA~\cite{han2024high} reconstructs relightable 3D face assets from a single co-located smartphone flashlight sequence captured in a dim room. Similarly, Azinovi\'c \etal~\cite{azinovic2023high} additionally attach polarization foils and capture two such sequences with perpendicular polarization orientation to separate skin surface and subsurface reflectance. Using a co-located light and camera setup, these methods assume the position of the dominant light source is known and no shadowing term needs to be modeled. Instead of using a smartphone flashlight, SunStage~\cite{wang2023sunstage} takes a selfie video rotating under the sun as input and uses the varying angles between the sun and the face as guidance. Cast shadows from the sun are modeled by shadow mapping. It also jointly optimizes the sun's position together with face geometry and appearance. All these methods assume the specular reflection (and the shadowing) of the face comes from a single dominant point or directional light source in the capture environment, and the ambient light contributes only to a low-frequency diffuse term. Another line of work~\cite{dib2021practical, dib2021towards, dib2022s2f2, lattas2020avatarme, han2023learning, papantoniou2023relightify, dib2024mosar} aims to reconstruct shape and reflectance properties from a single portrait image, often relying heavily on statistical shape and appearance priors~\cite{blanz2023morphable, gerig2018morphable, smith2020morphable} which limits its expressiveness. Our work instead focuses on accurate personalized reconstruction. Rainer \etal~\cite{rainer2023neural} use tiny shading networks to disentangle shading from explicit reflectance maps but assume a known high-quality geometry and smooth lighting. NeuFace~\cite{zheng2023neuface} represents the face with a neural SDF and proposes to learn appearance factorization under unknown low frequency light with a novel neural BRDF basis. However, it requires multi-view input similar to those from a light stage setup. Closest to ours, FLARE~\cite{bharadwaj2023flare} builds relightable head avatars from monocular videos. It adopts the split-sum approximation~\cite{karis2013real, Munkberg_2022_CVPR} for relighting and a neural version of it during training. This rendering model ignores the self-occlusion and bakes part of the shading into the albedos. In contrast, we propose a modified formulation of the split-sum approximation which explicitly handles light visibility and combine it with ray tracing, leading to higher quality shape and appearance reconstruction.

\section{Monocular Appearance Capture}
\label{sec:method}

We now describe our method for facial appearance capture given a monocular head rotation sequence. We assume the expression of the subject does not change throughout the sequence. As a pre-processing step, we run monocular tracking based on landmarks~\cite{chandran2023continuous} and a photometric loss~\cite{qian2024versatile} to obtain an initial canonical mesh from a 3DMM fit. We also estimate a fixed camera pose, per-frame rigid head poses and neck rotation. For more details about the pre-processing step please see the supplemental PDF. The output of our inverse rendering system will be a 3D asset of the subject containing a high-quality mesh, and diffuse albedo, specular intensity and roughness as 2D texture maps. In the following, we first describe our geometry optimization formulation
in Section~\ref{sec:geo}, our novel occlusion-aware shading model in Section~\ref{sec:shading}, and then optimization details in Section~\ref{sec:impl}.

\subsection{Geometry Optimization}
\label{sec:geo}

Balancing the updates to both geometry and textures in inverse rendering can be a challenging task. More specifically, the optimization might overfit too quickly on the textural components before learning the correct geometry. This is often the case when Laplacian shape regularization is applied to enforce geometric smoothness, making the geometry update too slow. Related work such as FLARE~\cite{bharadwaj2023flare} addresses this issue by using a two-stage approach. First, a detailed geometry and only blurry textures are learned, and then, the shape is fixed and textures are learned after re-initialization in the second stage.
Our method, however, optimizes geometry and textures at the same time. To do so, we adopt a preconditioning framework similar to Nicolet~\etal~\cite{nicolet2021large}, which biases gradient steps towards smooth solutions. The vertex positions $\mathbf{v}$ in each iteration are updated by
\begin{equation}
	\mathbf{v} \leftarrow \mathbf{v} - \eta (\mathbf{I} + \lambda_{\text{geo}} \mathbf{L})^2 \frac{\partial \mathcal{L}}{\partial \mathbf{v}},
\end{equation}
where $\mathbf{v} \in R^{N \times 3}$ collects mesh vertex positions along rows, $\eta$ is the learning rate, $\mathbf{I}$ is an identity matrix, $\mathbf{L}$ is the uniform Laplacian, and $\mathcal{L}$ is our loss function described in~\secref{sec:impl}. The hyper-parameter $\lambda_{geo} > 0$ balances between the original objective of matching the input images and a smooth mesh. We set $\lambda_{geo} = 19$ in all our experiments. This way, we can apply a large learning rate for the geometry optimization while keeping the mesh smooth and self-intersection free.

\subsection{Occlusion-Aware Shading Model}
\label{sec:shading}

Following the rendering equation~\cite{kajiya1986rendering}, we compute the outgoing radiance $L(\omega_o)$ at location $\mathbf{x}$ from direction $\omega_o$ by:
\begin{equation}
\label{eq:render}
	L(\omega_o) = \int_{\Omega}  f(\mathbf{x}, \omega_i, \omega_o) L_i(\omega_i) (\omega_i \cdot \mathbf{n}) d \omega_i.
\end{equation}
We decompose the BRDF $f(\mathbf{x}, \omega_i, \omega_o)$ into the sum of a diffuse term $f_d$ and a specular term $f_s$. We use the simple Lambertian model for the diffuse term:
\begin{equation}
	f_d(\mathbf{x}) = \frac{\rho(\mathbf{x})}{\pi},
\end{equation}
where $\rho$ is the diffuse albedo. We use a specular BRDF similar to Kelemen and Szirmay-Kalos~\cite{kelemen2001microfacet}, which has been shown to be well suited for rendering human skin~\cite[Chapter 14]{gpugems3}:
\begin{equation}
	f_s(\mathbf{x}, \omega_i, \omega_o) = \frac{DGF}{4(\omega_i \cdot \mathbf{n})(\omega_o \cdot \mathbf{n})},
\end{equation}
where $D$, $G$, and $F$ are functions representing the Beckmann normal distribution, geometric attenuation and Fresnel terms, respectively.

\paragraph{Accounting for self-occlusion.}
We consider direct illumination where $L_i(\omega_i)$ comes only from the light sources. Spherical Harmonics~\cite{wo1928spherical,Chen2019-xo} or Spherical Gaussians~\cite{Yan2012-cq} are popular representations used by prior methods~\cite{physg2021, zheng2023neuface} but they can only model low- to medium-frequency lighting. We therefore follow other work, \eg FLARE~\cite{bharadwaj2023flare}, and use a differentiable split-sum approximation~\cite{kautz2000unified, mcallister2002efficient, Munkberg_2022_CVPR}, which allows to capture all-frequency lighting.  Unfortunately this approach ignores self-shadowing, and so we propose a novel visibility-modulated split-sum approximation to account for self-shadowing, which we describe in the following.

We start by introducing the split-sum formulation from Karis~\cite{karis2013real} which approximates ~\eqnref{eq:render} as%
\begin{equation}
\label{eq:split-approx}
\begin{aligned}
& L(\omega_o) \approx \\
& \int_{\Omega} f(\mathbf{x}, \omega_i, \omega_o)(\omega_i \cdot \mathbf{n}) d\omega_i \int_{\Omega}L_i(\omega_i) D(\mathbf{h}) (\omega_i \cdot \mathbf{n}) d\omega_i.
\end{aligned}
\end{equation}

The first term is the integral of the BRDF under a solid white environment map, \ie $L_i(\omega_i) = 1, \forall \omega_i$. It only depends on $\cos \theta = \omega_i \cdot \mathbf{n}$ and the roughness $r$. Therefore, it can be precomputed and stored as a 2D look-up texture. The second integral is a pre-filtered environment map, where $D$ is the normal distribution function of the BRDF and $\mathbf{h} = \frac{\omega_i + \omega_o}{\|\omega_i + \omega_o\|_2}$ is the half vector. This term can also be precomputed as a mipmap by convolving the environment map with $D$ at different roughness values. At rendering time, the outgoing radiance can then be efficient computed using only two texture lookups. Note that when evaluating ~\eqnref{eq:split-approx}, the contribution of the prefiltered environment map is only dependent on the reflected view direction $\omega_r$ and the roughness $r$. This is not physically accurate as the contribution should be reduced if the light is (partially) blocked from the upper hemisphere at a specific point. To account for self-occlusion, we must modulate $L_i(\omega_i)$ differently at different locations based on the light visibility $V(\mathbf{x}, \omega_i)$, \ie, making the second integral in ~\eqnref{eq:split-approx} $ \int_{\Omega}L_i(\omega_i)V(\mathbf{x}, \omega_i)D(\mathbf{h}) (\omega_i \cdot \mathbf{n}) d\omega_i$. This new integral cannot be precomputed anymore since it's different for different $\mathbf{x}$. However, we notice that when $r \ll 1$, the new integral is 0 unless $\omega_i$ is close to $\omega_r$. Therefore, we can approximate it as

\begin{figure*}[t]
	\centering
	\includegraphics[trim=0 0 0 0, clip, width=0.95\linewidth]{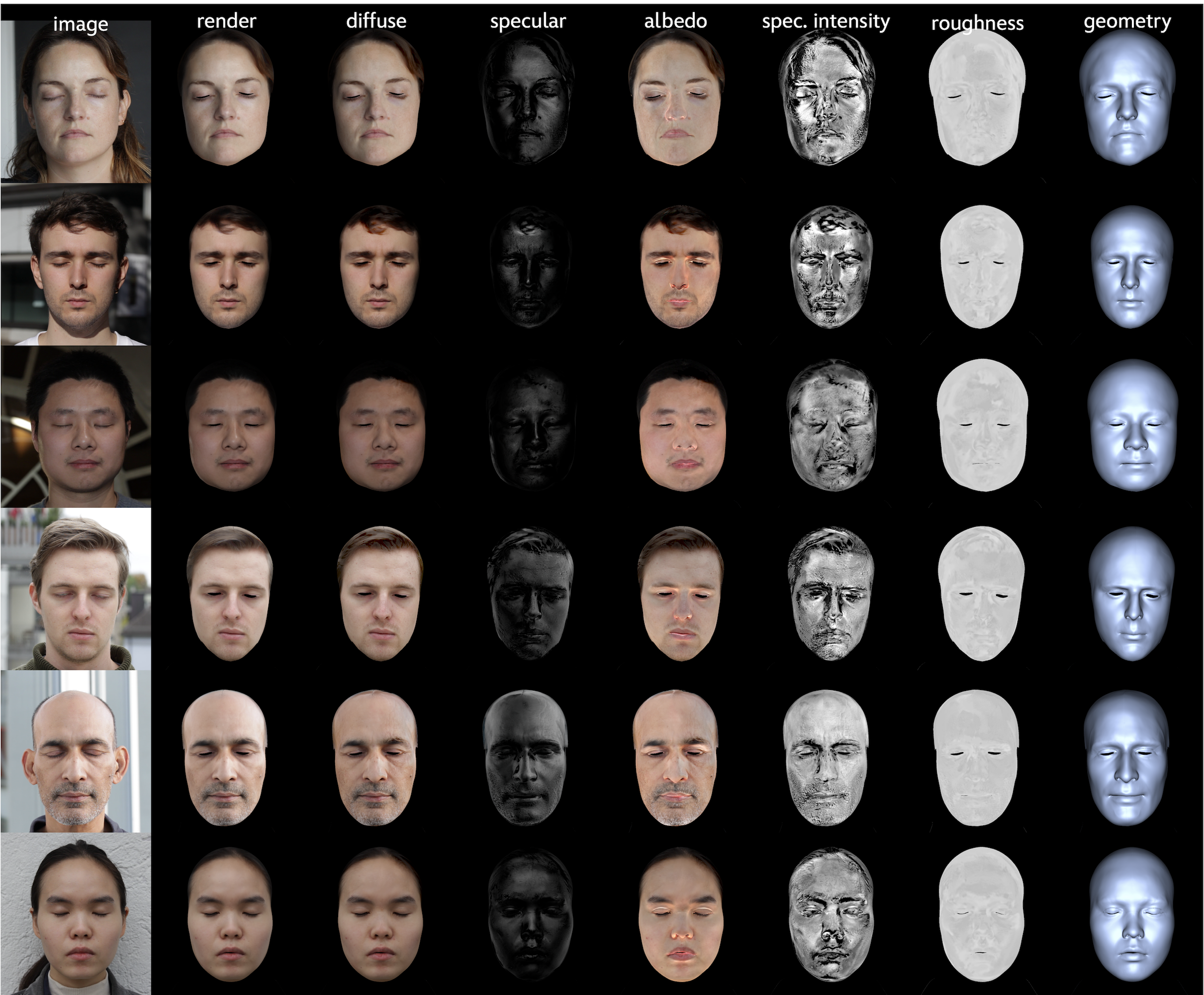}
	\caption{A selection of facial appearance reconstruction and decomposition results for different subjects in different environments, both indoors and outdoors, with sunny and cloudy sky.}
	\label{fig:hero}
\end{figure*}

\begin{equation}
\label{eq:vis-approx}
\begin{aligned}
& \int_{\Omega}L_i(\omega_i)V(\mathbf{x}, \omega_i)D(\mathbf{h}) (\omega_i \cdot \mathbf{n}) d\omega_i \\
\approx & \tilde{V}(\mathbf{x}, \omega_r) \int_{\Omega}L_i(\omega_i) D(\mathbf{h}) (\omega_i \cdot \mathbf{n}) d\omega_i.
\end{aligned}
\end{equation}
Here, $\tilde{V}(\mathbf{x}, \omega_r)$ is the view-dependent visibility. The simplest choice of $\tilde{V}(\mathbf{x}, \omega_r)$ is to only evaluate the light visibility at $\omega_r$, \ie, $\tilde{V}(\mathbf{x}, \omega_r) \coloneqq V(\mathbf{x}, \omega_r)$. Intuitively, this means the outgoing radiance is 0 if the reflected view direction is occluded. Note that this approximation is exact for perfect specular reflection (\ie mirrors). To add some softness in the visibility term (instead of considering it as a binary function), we approximate it by Monte Carlo integration,
\begin{equation}
\tilde{V}(\mathbf{x}, \omega_r) \coloneqq \frac{1}{K} \sum_{k=1}^{K} \frac{V(\mathbf{x}, \omega_k)}{D(\mathbf{n}, \omega_k, \omega_r, r)},
\end{equation}
where the samples are drawn following the normal distribution of the BRDF. This bears some resemblance to the appearance models of Saito~\etal~\cite{saito2024relightable}. However, they parameterize the view-dependent specular visibility term using a neural network as they work with a large amount of studio data where the lighting is controlled and calibrated. Note however, ~\eqnref{eq:vis-approx} introduces large errors for rough surfaces (\ie, when $r \ll 1$ does not hold). We therefore use it only for the specular component, and implemented ray tracing with multiple importance sampling~\cite{veach1995optimally} for the diffuse component using the OptiX~\cite{parker2010optix} engine.

\subsection{Optimization Details}
\label{sec:impl}
Our reconstruction loss consists of an $L_1$ data term $\mathcal{L}_{\text{img}}$ and some regularization terms. We employ an $L_1$ mask loss $\mathcal{L}_{\text{mask}}$ between the mask obtained from MODNet~\cite{ke2022modnet} and the predicted binary mask. Although we use a parameterization similar to Nicolet \etal~\cite{nicolet2021large}, we find it helpful to still employ a Laplacian regularizer to stabilize the geometry optimization such that we do not need to set different $\lambda_{\text{geo}}$ for different datasets. This regularizer encourages the Laplacian of the optimized mesh to stay close to the Laplacian of the initial 3DMM fit
\begin{equation}
\mathcal{L}_{\text{Lap}} = \big\|\mathbf{L}(\mathbf{v} - \mathbf{v}_{\text{init}}) \big\|^2_2.
\end{equation}
We apply a white light regularization $\mathcal{L}_{\text{light}}$ on the environment map as in~\cite{Munkberg_2022_CVPR} and the roughness texture is regularized to be smooth via a total variation loss $\mathcal{L}_{\text{rough}}$. We noticed in our experiments that part of the specular signal tends to be baked into the diffuse albedo. We thus apply a weak regularization to encourage the diffuse render $I_{\text{diffuse}}$ to be small if possible, as
\begin{equation}
\mathcal{L}_{\text{diffuse}} = \big\|I_{\text{diffuse}}\big\|_2^2.
\end{equation}
The final loss is then
\begin{equation}
\begin{aligned}
\mathcal{L} \coloneqq & \mathcal{L}_{\text{img}} + \lambda_{\text{mask}} \mathcal{L}_{\text{mask}} + \lambda_{\text{Lap}} \mathcal{L}_{\text{Lap}} +\\
& \lambda_{\text{light}} \mathcal{L}_{\text{light}} + \lambda_{\text{rough}} \mathcal{L}_{\text{rough}} + \lambda_{\text{diffuse}} \mathcal{L}_{\text{diffuse}}.
\end{aligned}
\label{eq:loss}
\end{equation}

\section{Experiments}
\label{sec:experiments}

We now show several results of our appearance capture method, evaluate its performance compared to previous work, and offer several ablation studies to validate our design choices.  Please refer to our supplemental material for additional results.

\subsection{Appearance and Geometry Reconstruction}

\begin{figure}[!ht]
	\centering
	\includegraphics[trim=0 0 0 0, clip, width=\linewidth]{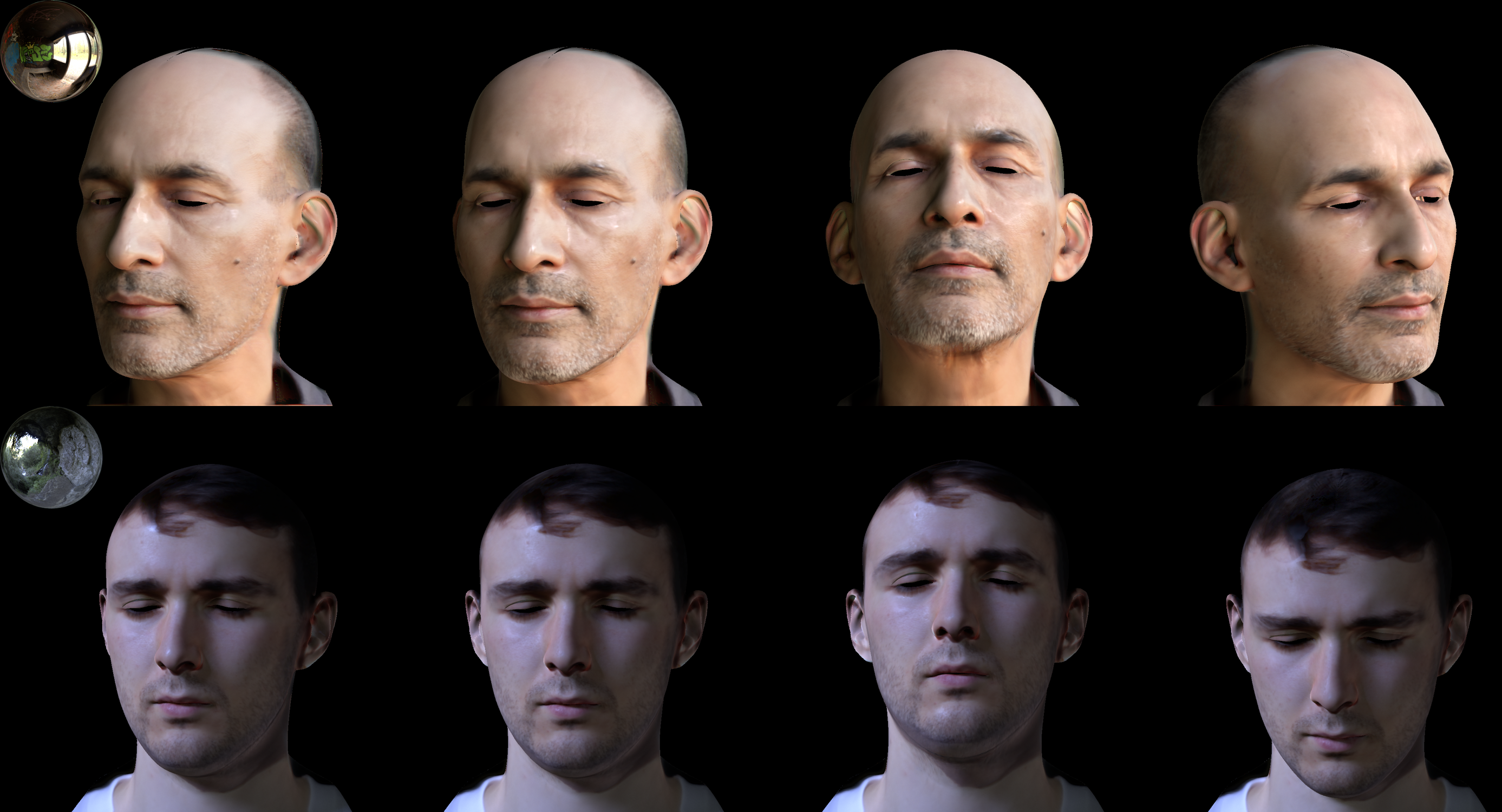}
	\caption{Relighting results of multiple frames on two different subjects in two different environments. The environment is shown top left of each row. See~\figref{fig:hero} for the original environment.}
	\label{fig:relit}
\end{figure}

We begin by highlighting the versatility of our approach by showing several appearance capture results of different subjects in different environments in~\figref{fig:hero}.  Each row of the figure illustrates one of the input images, the corresponding render using our recovered appearance properties, and then a breakdown of the reconstructed appearance maps (diffuse albedo, specular intensity, specular roughness) and geometry.  Our method can be applied indoors or outdoors, with sunny or cloudy skies. We show how the recovered geometry and appearance maps can be used to render the subject in a new environment by relighting them.  \figref{fig:relit} shows two subjects relit in two different environments (see~\figref{fig:hero} for the original environment). Please refer to the supplemental video for more results.

\subsection{Comparisons}

We compare our results to related methods for facial appearance capture in the wild: FLARE~\cite{bharadwaj2023flare}, NextFace~\cite{dib2021practical}, and SunStage~\cite{wang2023sunstage}.

\begin{figure*}[!ht]
	\centering
	\includegraphics[trim=0 0 0 0, clip, width=0.95\linewidth]{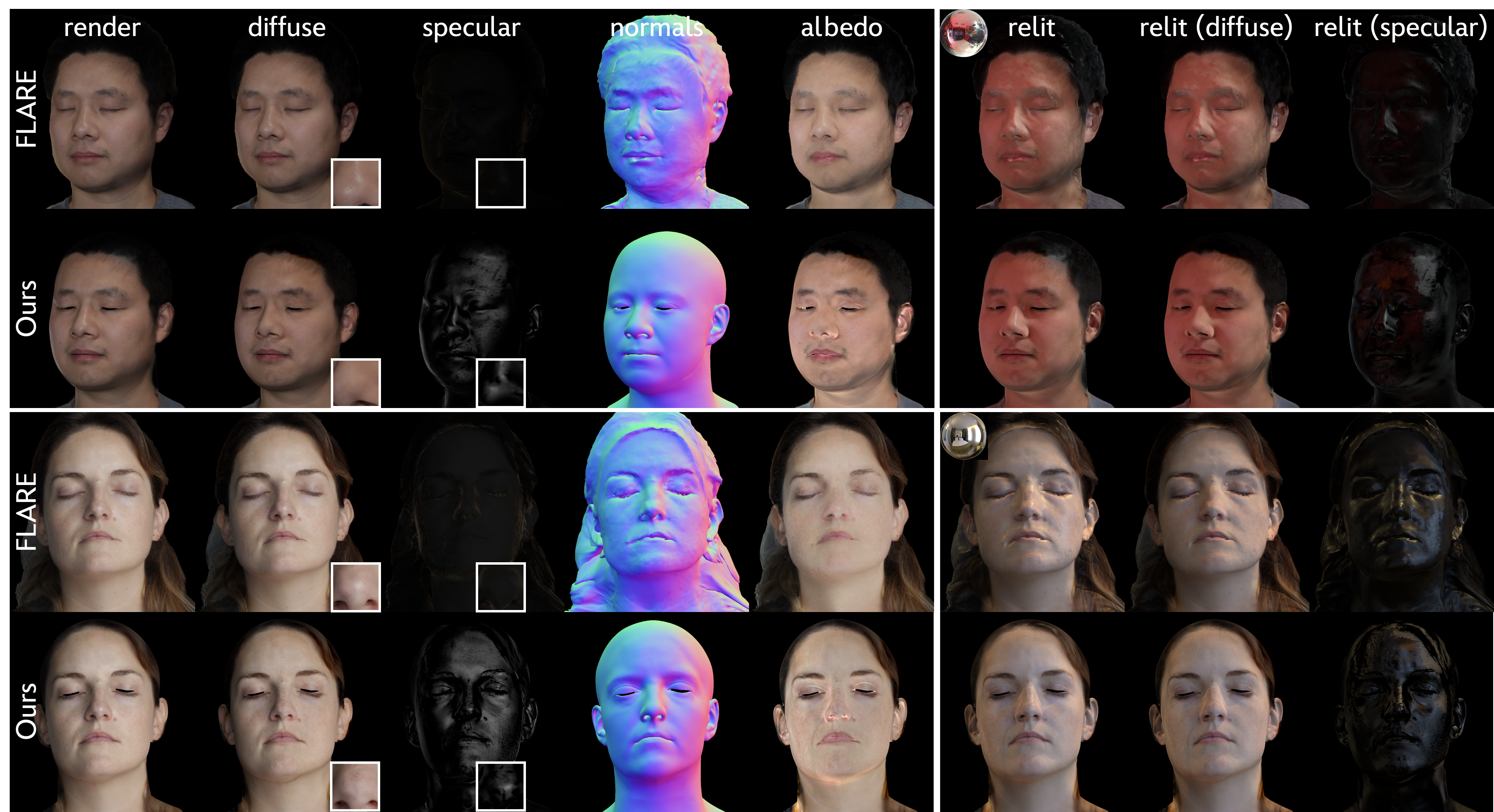}
	\caption{Qualitative comparisons with FLARE~\cite{bharadwaj2023flare}. We show one training frame on the left, with separated diffuse, specular, normals and diffuse albedo renders.  On the right we show a relit example using the reconstructed appearance.  FLARE fails to separate the diffuse and specular signals, produces noisy normals, and bakes shading into the diffuse albedo, all leading to poor results under relighting.
}
	\label{fig:flare}
\end{figure*}

\figref{fig:flare} shows a qualitative comparison of our method to FLARE on two different subjects.  While the combined final render is similar between FLARE and our method, it is clear that FLARE fails to separate the diffuse and specular components, baking most of the specular signal in the diffuse map resulting in a nearly zero specular render (\figref{fig:flare}, 3rd column).  In contrast, our shading model is completely physically-based and correctly separates the diffuse and specular components.  The normals reconstructed by FLARE portray a lot of spatial noise and the geometry contains self-intersections, unlike ours (4th column).  The diffuse albedo from FLARE contains more baked-in diffuse shading than our result (5th column).  As a result, FLARE performs worse when relighting the subject under a novel environment map than our approach (columns 6, 7 and 8). 

\begin{figure*}[!ht]
	\centering
	\includegraphics[trim=0 0 0 0, clip, width=0.9\linewidth]{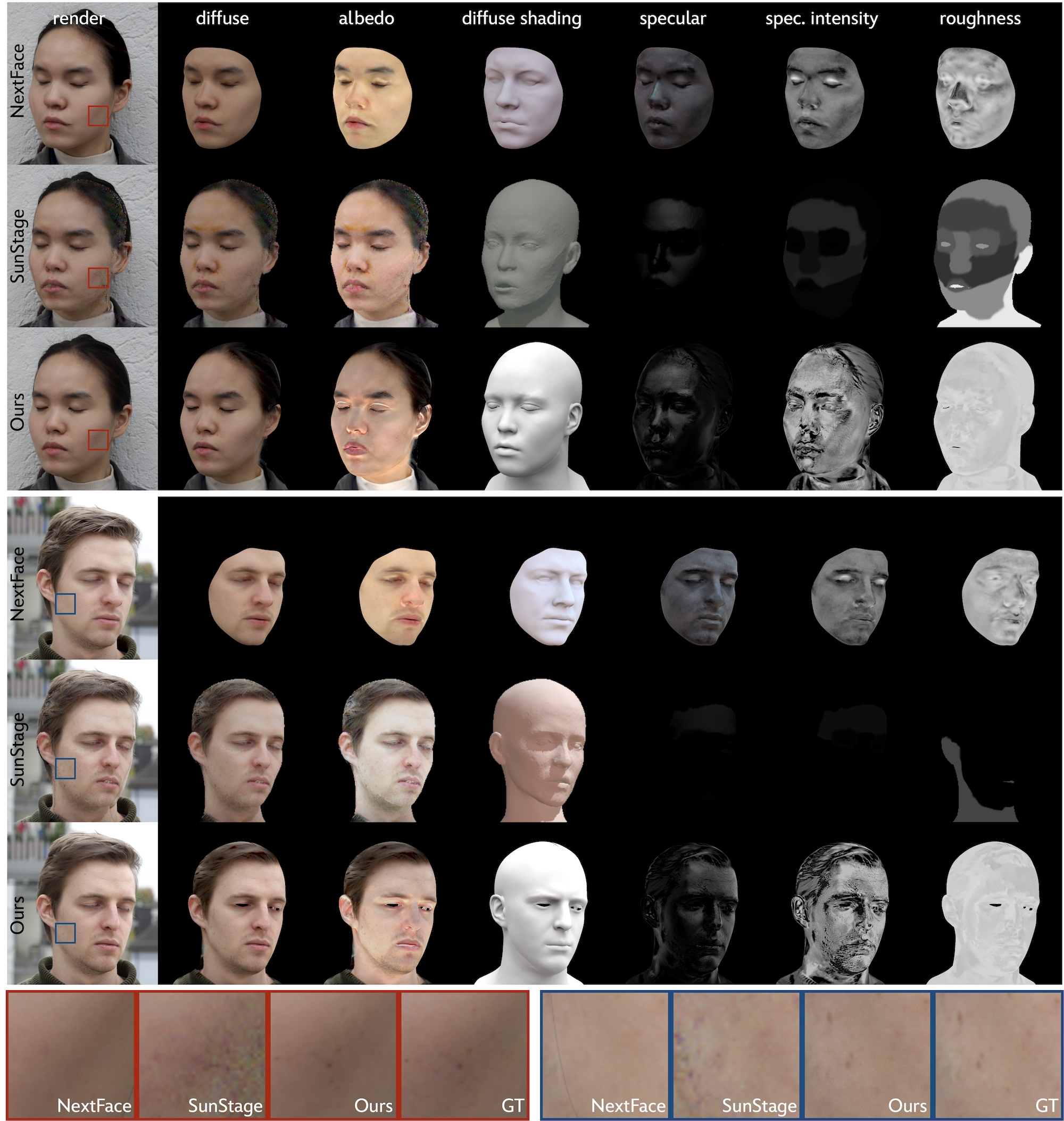}
	\caption{Qualitative comparisons with NextFace~\cite{dib2021practical} and SunStage~\cite{wang2023sunstage} on two different subjects. The first column is the resulting render overlaid on one input image, and the remaining columns indicate the recovered appearance maps.  NextFace produces inaccurate shapes and blurry appearance, where SunStage produces poor shadows and incorrect diffuse/specular separation.  Our method produces the most accurate results, also indicated by the zoom region in the final row.}	\label{fig:nextface-sunstage}
\vspace{-5mm}
\end{figure*}

We also perform qualitative comparisons to NextFace and SunStage in \figref{fig:nextface-sunstage}. NextFace relies on a statistical prior, leading to inaccurate shapes and blurry appearance for different subjects (\figref{fig:nextface-sunstage}, rows 1 and 4). It produces very similar shapes for the two subjects shown in~\figref{fig:nextface-sunstage}, while our geometry preserves the identity and likeness of the subjects. SunStage assumes a single point light in the scene, and the shadows and specular components come only from this point light. This leads to poor diffuse and specular separation and incorrect shadows in generic environments like the examples shown here (\figref{fig:nextface-sunstage}, rows 2 and 5). In the first subject, we see orange artifacts on the forehead of the diffuse albedo in the SunStage result (row 2). In the second subject, the specular component is completely missing (row 5). The shape and textures from SunStage also have lower resolution than ours.  The final row of \figref{fig:nextface-sunstage} illustrates the appearance details for zoomed-in regions shown by the red and blue squares, indicating that our method produces the most accurate details.

As a quantitative comparison, we show reconstruction errors of to NextFace, SunStage and FLARE compared to ours in \tabref{tab:quantitative}.  For fairness, we compute errors only on skin regions and average over all subjects in the dataset.  Our method prevails in all metrics. The metrics are computed in linear RGB space and averaged over all tested subjects.

\begin{table}
	\centering
	\footnotesize
	\begin{tabular}{l|cccc}
		\toprule
		& PSNR $\uparrow$ & MAE $\downarrow$ & SSIM \cite{wang2004image} $\uparrow$ & LPIPS \cite{zhang2018unreasonable} $\downarrow$ \\
		\midrule
		NextFace \cite{dib2021practical} & 25.30 & 10.63 &  0.78 & 0.31\\
		SunStage  \cite{wang2023sunstage} & 29.47 & 5.28 & 0.88 & 0.14 \\
		FLARE  \cite{bharadwaj2023flare} & 30.40 & 2.01 & 0.94 & 0.15 \\
		Ours (w/o vis) & 34.55 &  1.79 &  0.96 &  \textbf{0.10} \\
		\midrule
	    Ours & \textbf{38.09} &  \textbf{1.18} &  \textbf{0.97} &  \textbf{0.10} \\
		\bottomrule
	\end{tabular}
	\caption{Reconstruction errors computed over the skin region averaged for all the subjects. }
	\label{tab:quantitative}
\end{table}

\subsection{Evaluations}

\begin{figure}[tb]
	\centering
	\includegraphics[trim=0 0 0 0, clip, width=\linewidth]{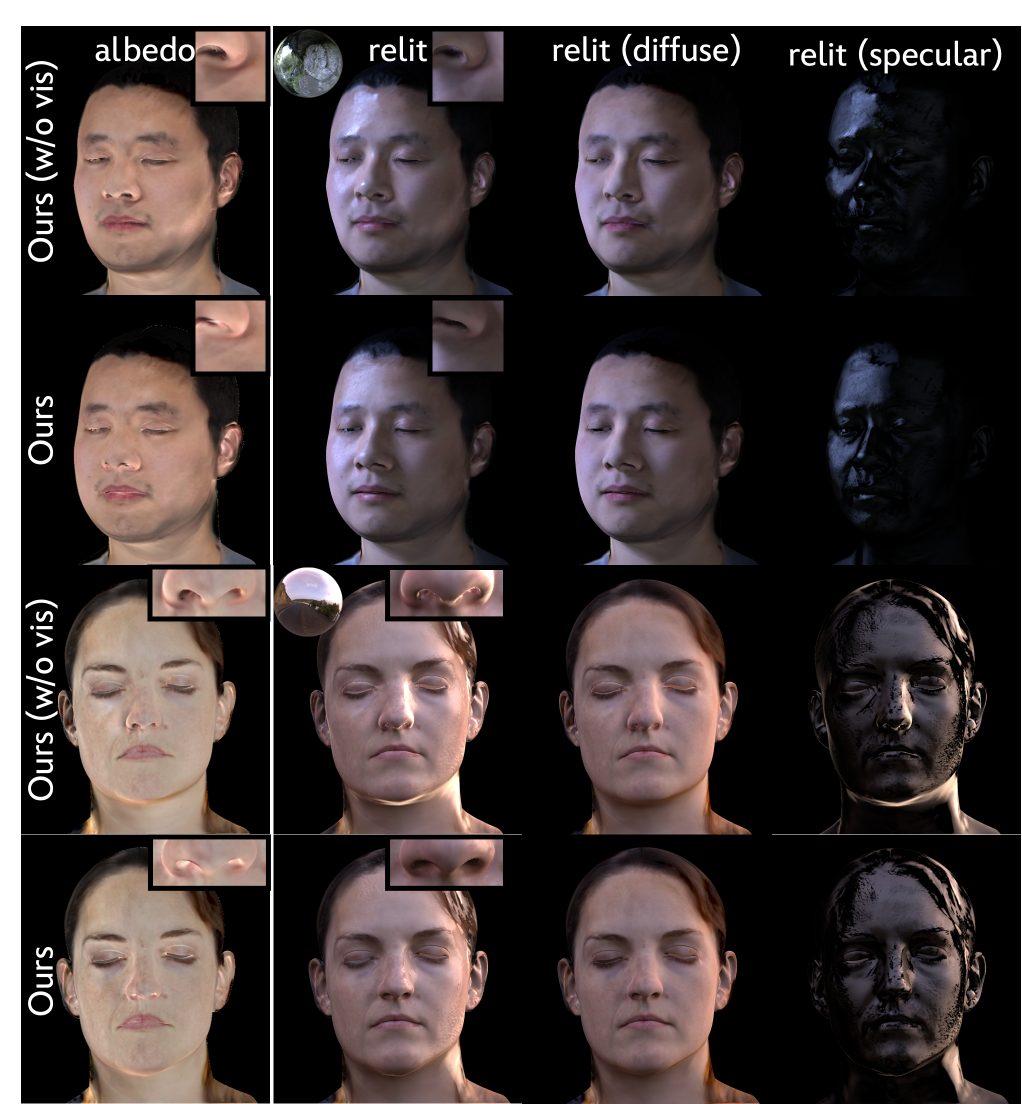}
	\caption{Relighting evaluation of our method with and without accounting for self-occlusion. We also visualize the corresponding diffuse albedo, diffuse render and specular render.}
	\label{fig:relit-ablation}
\vspace{-3mm}
\end{figure}

\paragraph{Effects of light visibility.} 
We first evaluate the effects of accounting for self-occlusion by comparing the relit renders of our method with the original split-sum approximation (denoted as \textit{Ours (w/o vis)}) in~\figref{fig:relit-ablation}. We can see that when self-occlusion is not accounted for, shadows under the capture lighting got baked into the albedo (column 1 with zoomed-in patches focusing around the nose), leading to wrong shadows when relit (row 1, column 2, 3). The specular component also exhibits unrealistic sharp highlight (row 1, column 2, 4). Moreover, when the major light source is behind the subject, the baseline renders show artifacts in the form of strong specular highlights on the side of the nose and under the chin (row 3, column 2, 4) since self-occlusion is not properly handled. In contrast, our proposed model correctly removes baked-in shading and produces more realistic relit renders (row 2, 4). We also quantitatively evaluated the performance gain of accounting for self-occlusion in terms of reconstruction errors in~\tabref{tab:quantitative}. 

\begin{figure}[tb]
	\centering
	\includegraphics[trim=0 0 0 0, clip, width=\linewidth]{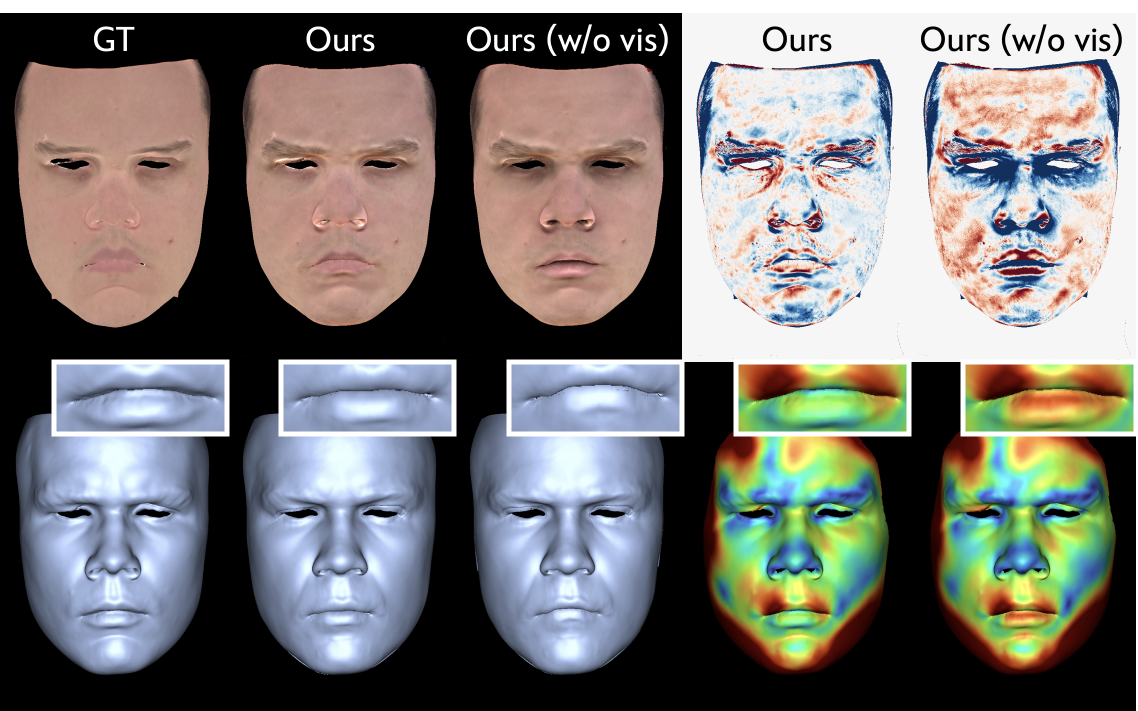}
	\caption{Comparing our method with and without accounting for light visibility on a synthetic dataset. The mesh errors are displayed with a scale of 0mm~\includegraphics[width=2cm]{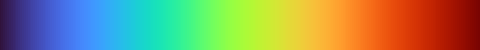}~5mm and the albedo error with a scale of -0.1~\includegraphics[width=2cm]{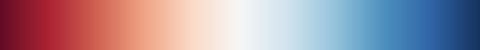}~0.1.}
	\label{fig:raytrace}
\vspace{-3mm}
\end{figure}

\paragraph{Evaluation on synthetic data.}
To better evaluate the quality of the reconstructed mesh and textures against ground truth, we created a synthetic dataset with a monocular head rotation sequence similar to the real data. The ground truth mesh and albedo are from a studio appearance capture method \cite{riviere2020single} (Fig.~\ref{fig:raytrace} column 1). We use a natural outdoor environment map on a cloudy day as lighting. Please refer to the supplementary document for more details on the synthetic dataset. We show reconstructed diffuse albedo and geometry of our method with and without light visbility accounted for in columns 2 and 3 of Fig.~\ref{fig:raytrace}, and the error maps compared with the ground truth are shown in the last two columns. We can see that our method produces a more accurate reconstruction of the diffuse albedo while the baseline result contains a lot of baked-in shading. In terms of shape reconstruction, the two perform similarly. However, the baseline method does not handle shadows well, leading to slightly worse shape recovery in regions where self-occlusion plays an important role, \eg, the lips. 

\begin{figure}[tb]
	\centering
	\includegraphics[trim=0 0 0 0, clip, width=\linewidth]{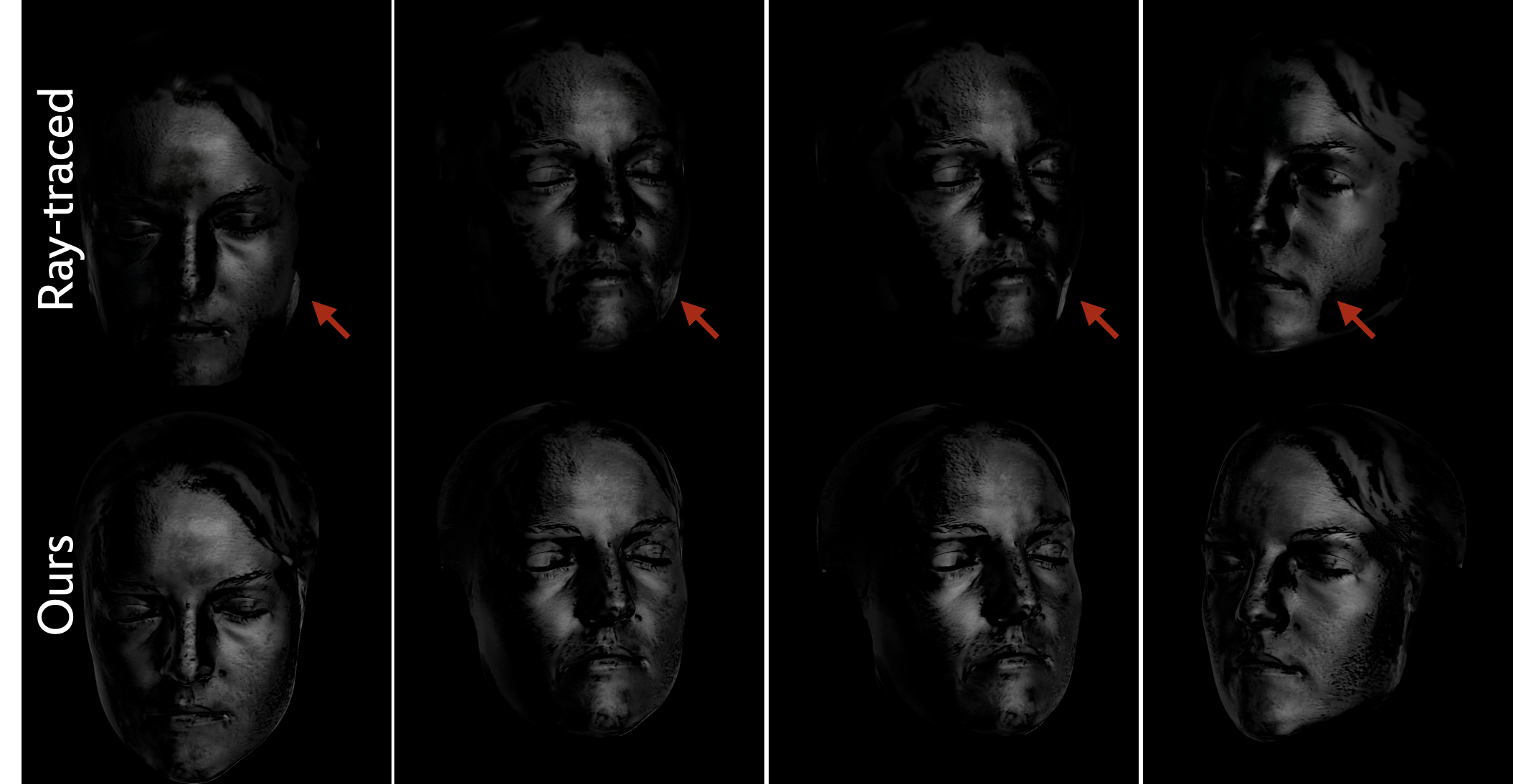}
	\caption{Comparing the specular renders from our method and a ray-tracer across frames in the same head rotation sequence.}
	\label{fig:raytrace-spec}
\vspace{-3mm}
\end{figure}

\vspace{-3mm}
\paragraph{Ours \textit{vs} raytracing for specular.}
While a ray-tracer handles visibility inherently, we find that our modified split-sum approximation behaves more stably in our inverse rendering setting when we need to solve for shape, textures, lighting at the same time from merely monocular input. Ray tracing produces flickery specular images (Fig.~\ref{fig:raytrace-spec} row 2) where part of the skin abruptly changes from very bright to very dark while the head is rotating. We also notice high frequency artifacts, \eg sharp boundary between bright and dark pixels, in the ray traced render, as denoted by the red arrows. In contrast, our model gives visually smoother specular renders (Fig.~\ref{fig:raytrace-spec}, row 3). Please refer to the supplemental videos for an animated visualization.

\begin{figure}[tb]
	\centering
	\includegraphics[trim=0 0 0 0, clip, width=\linewidth]{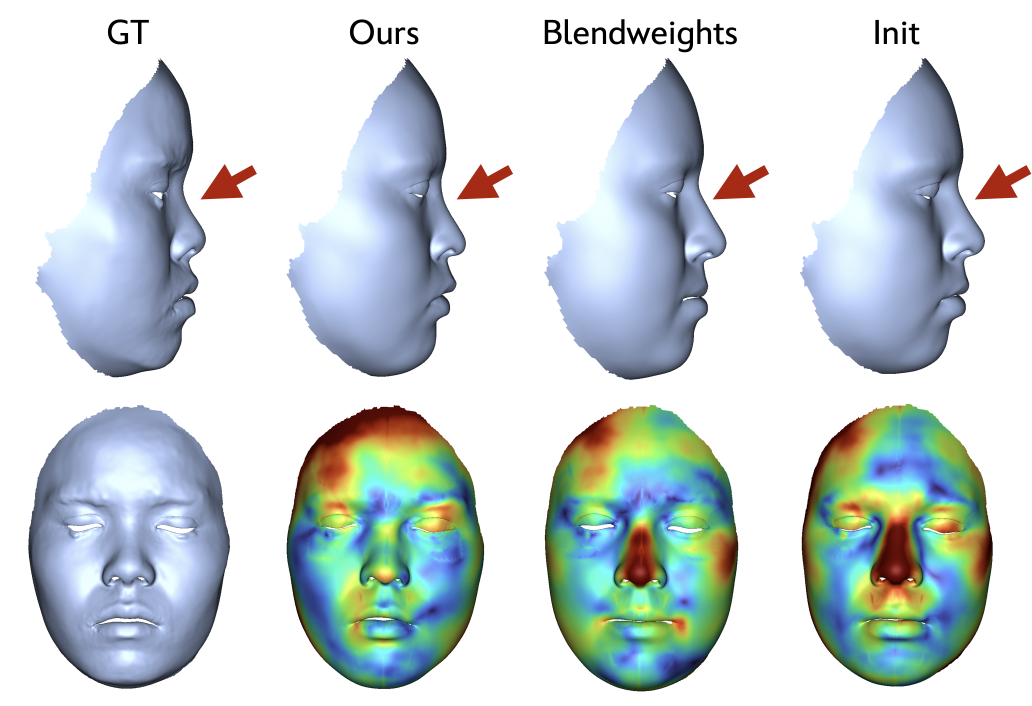}
	\caption{Our geometry optimization pipeline recovers the correct nose shape from shading, achieving the better reconstruction quality compared to 3DMM fit. The mesh errors are displayed with a scale of 0mm~\includegraphics[width=2cm]{figures/colormap.png}~5mm.}
	\label{fig:blendweights}
\end{figure}

\vspace{-4mm}
\paragraph{Optimizing vertex positions \textit{vs} blendweights.}
We choose to optimize vertex positions directly instead of parameterizing the shape using blendweights of a 3DMM as in NextFace \cite{dib2021practical}. We show in Fig.~\ref{fig:blendweights} that our model achieves the least reconstruction error in the central part of the face compared to the initial mesh or the one optimized through a 3DMM in the same inverse rendering setting. Even though the silhouette of the side face is never shown in the training data, our method recovers the correct shape of the nose from shading. The ground truth reference mesh is a 3D facial scan from a multi-view face scanner \cite{beeler2010high}.

\begin{figure}[tb]
	\centering
	\includegraphics[trim=0 0 0 0, clip, width=\linewidth]{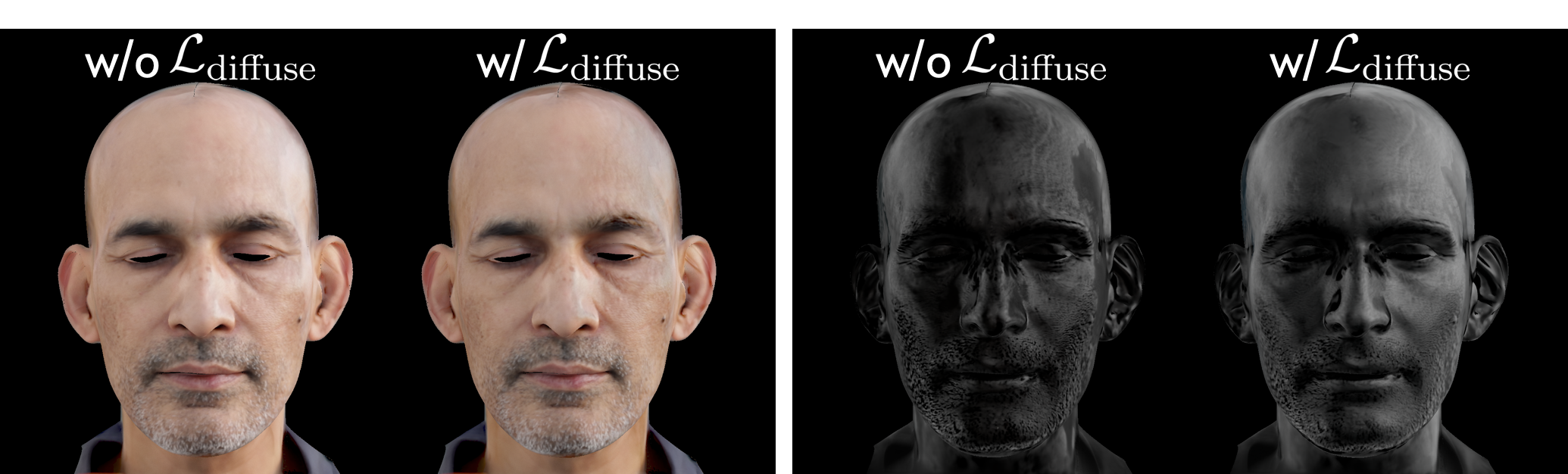}
	\caption{Diffuse and specular component without and with the regularization term $\mathcal{L}_{\text{diffuse}}$.}
	\label{fig:diffuse-reg}
	\vspace{-3mm}
\end{figure}

\vspace{-4mm}
\paragraph{Regularize diffuse component.}
Last, we apply a weak regularization to encourage the diffuse render to be small which prevents too much specular from being baked into the diffuse component, as we show in Fig.~\ref{fig:diffuse-reg}. This gives us better separation of the diffuse and specular signals.

\section{Limitations}
\label{sec:limitation}
We assume the head poses (and neck rotations) are provided as input to our method. Inaccuracies in the head pose estimation impair the reconstruction quality of our method substantially (see failure cases in the supplemental). 
Although we make no assumptions on the lighting condition, we cannot recover the appearance if part of the face is in extreme shadows in all frames. Our current face template geometry does not model eyes, however switching to a different template model would allow to reconstruct eyes better. Also, correct skin tone recovery is not guaranteed due to the ambiguity between the illumination and appearance.

\section{Conclusion}
\label{sec:conclusion}
We propose a new lightweight facial appearance capture method, surpassing the quality of existing lightweight approaches. It works truly in the wild, making no assumption on the environment illumination. Our novel shading model explicitly accounts for self-occlusion, leading to faithful recovery of the shape and appearance properties from a monocular video containing a simple head rotation.
{
    \small
    \bibliographystyle{ieeenat_fullname}
    \bibliography{main}

\begin{thebibliography}{74}
\providecommand{\natexlab}[1]{#1}
\providecommand{\url}[1]{\texttt{#1}}
\expandafter\ifx\csname urlstyle\endcsname\relax
  \providecommand{\doi}[1]{doi: #1}\else
  \providecommand{\doi}{doi: \begingroup \urlstyle{rm}\Url}\fi

\bibitem[Azinovi{\'c} et~al.(2023)Azinovi{\'c}, Maury, Hery, Nie{\ss}ner, and
  Thies]{azinovic2023high}
Dejan Azinovi{\'c}, Olivier Maury, Christophe Hery, Matthias Nie{\ss}ner, and
  Justus Thies.
\newblock High-res facial appearance capture from polarized smartphone images.
\newblock In \emph{Proceedings of the IEEE/CVF Conference on Computer Vision
  and Pattern Recognition}, pages 16836--16846, 2023.

\bibitem[Baert et~al.(2024)Baert, Bharadwaj, Castan, Maujean, Christie,
  Abrevaya, and Boukhayma]{baert2024spark}
Kelian Baert, Shrisha Bharadwaj, Fabien Castan, Benoit Maujean, Marc Christie,
  Victoria Abrevaya, and Adnane Boukhayma.
\newblock Spark: Self-supervised personalized real-time monocular face capture.
\newblock \emph{arXiv preprint arXiv:2409.07984}, 2024.

\bibitem[Beeler et~al.(2010)Beeler, Bickel, Beardsley, Sumner, and
  Gross]{beeler2010high}
Thabo Beeler, Bernd Bickel, Paul Beardsley, Bob Sumner, and Markus Gross.
\newblock High-quality single-shot capture of facial geometry.
\newblock In \emph{ACM SIGGRAPH 2010 papers}, pages 1--9. 2010.

\bibitem[Bharadwaj et~al.(2023)Bharadwaj, Zheng, Hilliges, Black, and
  Fernandez-Abrevaya]{bharadwaj2023flare}
Shrisha Bharadwaj, Yufeng Zheng, Otmar Hilliges, Michael~J Black, and Victoria
  Fernandez-Abrevaya.
\newblock Flare: Fast learning of animatable and relightable mesh avatars.
\newblock \emph{arXiv preprint arXiv:2310.17519}, 2023.

\bibitem[Bi et~al.(2024)Bi, Zeng, Zeng, Pei, Feng, Zhou, and Wu]{bi2024rgs}
Zoubin Bi, Yixin Zeng, Chong Zeng, Fan Pei, Xiang Feng, Kun Zhou, and Hongzhi
  Wu.
\newblock Gs\textsuperscript{3}: Efficient relighting with triple gaussian
  splatting.
\newblock In \emph{SIGGRAPH Asia 2024 Conference Papers}, 2024.

\bibitem[Blanz and Vetter(2023)]{blanz2023morphable}
Volker Blanz and Thomas Vetter.
\newblock A morphable model for the synthesis of 3d faces.
\newblock In \emph{Seminal Graphics Papers: Pushing the Boundaries, Volume 2},
  pages 157--164. 2023.

\bibitem[Boss et~al.(2021{\natexlab{a}})Boss, Braun, Jampani, Barron, Liu, and
  Lensch]{boss2021nerd}
Mark Boss, Raphael Braun, Varun Jampani, Jonathan~T Barron, Ce Liu, and Hendrik
  Lensch.
\newblock Nerd: Neural reflectance decomposition from image collections.
\newblock In \emph{Proceedings of the IEEE/CVF International Conference on
  Computer Vision}, pages 12684--12694, 2021{\natexlab{a}}.

\bibitem[Boss et~al.(2021{\natexlab{b}})Boss, Jampani, Braun, Liu, Barron, and
  Lensch]{boss2021neural}
Mark Boss, Varun Jampani, Raphael Braun, Ce Liu, Jonathan Barron, and Hendrik
  Lensch.
\newblock Neural-pil: Neural pre-integrated lighting for reflectance
  decomposition.
\newblock \emph{Advances in Neural Information Processing Systems},
  34:\penalty0 10691--10704, 2021{\natexlab{b}}.

\bibitem[Boss et~al.(2022)Boss, Engelhardt, Kar, Li, Sun, Barron, Lensch, and
  Jampani]{boss2022samurai}
Mark Boss, Andreas Engelhardt, Abhishek Kar, Yuanzhen Li, Deqing Sun, Jonathan
  Barron, Hendrik Lensch, and Varun Jampani.
\newblock Samurai: Shape and material from unconstrained real-world arbitrary
  image collections.
\newblock \emph{Advances in Neural Information Processing Systems},
  35:\penalty0 26389--26403, 2022.

\bibitem[Chandran et~al.(2020)Chandran, Bradley, Gross, and
  Beeler]{chandran2020semantic}
Prashanth Chandran, Derek Bradley, Markus Gross, and Thabo Beeler.
\newblock Semantic deep face models.
\newblock In \emph{2020 international conference on 3D vision (3DV)}, pages
  345--354. IEEE, 2020.

\bibitem[Chandran et~al.(2023)Chandran, Zoss, Gotardo, and
  Bradley]{chandran2023continuous}
Prashanth Chandran, Gaspard Zoss, Paulo Gotardo, and Derek Bradley.
\newblock Continuous landmark detection with 3d queries.
\newblock In \emph{Proceedings of the IEEE/CVF Conference on Computer Vision
  and Pattern Recognition}, pages 16858--16867, 2023.

\bibitem[Chen et~al.(2019)Chen, Ling, Gao, Smith, Lehtinen, Jacobson, and
  Fidler]{Chen2019-xo}
Wenzheng Chen, Huan Ling, Jun Gao, Edward Smith, Jaakko Lehtinen, Alec
  Jacobson, and Sanja Fidler.
\newblock Learning to predict {3d} objects with an interpolation-based
  differentiable renderer.
\newblock \emph{Adv. Neural Inf. Process. Syst.}, 32, 2019.

\bibitem[Choi et~al.(2024)Choi, Yoon, Nam, Lee, and
  Baek]{choi2024differentiable}
Seokjun Choi, Seungwoo Yoon, Giljoo Nam, Seungyong Lee, and Seung-Hwan Baek.
\newblock Differentiable display photometric stereo.
\newblock In \emph{Proceedings of the IEEE/CVF Conference on Computer Vision
  and Pattern Recognition}, pages 11831--11840, 2024.

\bibitem[Debevec et~al.(2000)Debevec, Hawkins, Tchou, Duiker, Sarokin, and
  Sagar]{debevec2000acquiring}
Paul Debevec, Tim Hawkins, Chris Tchou, Haarm-Pieter Duiker, Westley Sarokin,
  and Mark Sagar.
\newblock Acquiring the reflectance field of a human face.
\newblock In \emph{Proceedings of the 27th annual conference on Computer
  graphics and interactive techniques}, pages 145--156, 2000.

\bibitem[Dib et~al.(2021{\natexlab{a}})Dib, Bharaj, Ahn, Th{\'e}bault,
  Gosselin, Romeo, and Chevallier]{dib2021practical}
Abdallah Dib, Gaurav Bharaj, Junghyun Ahn, C{\'e}dric Th{\'e}bault, Philippe
  Gosselin, Marco Romeo, and Louis Chevallier.
\newblock Practical face reconstruction via differentiable ray tracing.
\newblock In \emph{Computer Graphics Forum}, pages 153--164. Wiley Online
  Library, 2021{\natexlab{a}}.

\bibitem[Dib et~al.(2021{\natexlab{b}})Dib, Thebault, Ahn, Gosselin, Theobalt,
  and Chevallier]{dib2021towards}
Abdallah Dib, Cedric Thebault, Junghyun Ahn, Philippe-Henri Gosselin, Christian
  Theobalt, and Louis Chevallier.
\newblock Towards high fidelity monocular face reconstruction with rich
  reflectance using self-supervised learning and ray tracing.
\newblock In \emph{Proceedings of the IEEE/CVF International Conference on
  Computer Vision}, pages 12819--12829, 2021{\natexlab{b}}.

\bibitem[Dib et~al.(2022)Dib, Ahn, Thebault, Gosselin, and
  Chevallier]{dib2022s2f2}
Abdallah Dib, Junghyun Ahn, Cedric Thebault, Philippe-Henri Gosselin, and Louis
  Chevallier.
\newblock S2f2: Self-supervised high fidelity face reconstruction from
  monocular image.
\newblock \emph{arXiv preprint arXiv:2203.07732}, 2022.

\bibitem[Dib et~al.(2024)Dib, Hafemann, Got, Anderson, Fadaeinejad, Cruz, and
  Carbonneau]{dib2024mosar}
Abdallah Dib, Luiz~Gustavo Hafemann, Emeline Got, Trevor Anderson, Amin
  Fadaeinejad, Rafael~MO Cruz, and Marc-Andr{\'e} Carbonneau.
\newblock Mosar: Monocular semi-supervised model for avatar reconstruction
  using differentiable shading.
\newblock In \emph{Proceedings of the IEEE/CVF Conference on Computer Vision
  and Pattern Recognition}, pages 1770--1780, 2024.

\bibitem[Diederik(2014)]{diederik2014adam}
P~Kingma Diederik.
\newblock Adam: A method for stochastic optimization.
\newblock \emph{(No Title)}, 2014.

\bibitem[Engelhardt et~al.(2024)Engelhardt, Raj, Boss, Zhang, Kar, Li, Sun,
  Brualla, Barron, Lensch, et~al.]{engelhardt2024shinobi}
Andreas Engelhardt, Amit Raj, Mark Boss, Yunzhi Zhang, Abhishek Kar, Yuanzhen
  Li, Deqing Sun, Ricardo~Martin Brualla, Jonathan~T Barron, Hendrik Lensch,
  et~al.
\newblock Shinobi: Shape and illumination using neural object decomposition via
  brdf optimization in-the-wild.
\newblock In \emph{Proceedings of the IEEE/CVF Conference on Computer Vision
  and Pattern Recognition}, pages 19636--19646, 2024.

\bibitem[Enyo and Nishino(2024)]{Yenyo2024}
Yuto Enyo and Ko Nishino.
\newblock Diffusion reflectance map: Single-image stochastic inverse rendering
  of illumination and reflectance.
\newblock In \emph{Proceedings of the IEEE/CVF Conference on Computer Vision
  and Pattern Recognition (CVPR)}, 2024.

\bibitem[Feng et~al.(2021)Feng, Feng, Black, and Bolkart]{feng2021learning}
Yao Feng, Haiwen Feng, Michael~J Black, and Timo Bolkart.
\newblock Learning an animatable detailed 3d face model from in-the-wild
  images.
\newblock \emph{ACM Transactions on Graphics (ToG)}, 40\penalty0 (4):\penalty0
  1--13, 2021.

\bibitem[Gao et~al.(2023)Gao, Gu, Lin, Zhu, Cao, Zhang, and
  Yao]{gao2023relightable}
Jian Gao, Chun Gu, Youtian Lin, Hao Zhu, Xun Cao, Li Zhang, and Yao Yao.
\newblock Relightable 3d gaussian: Real-time point cloud relighting with brdf
  decomposition and ray tracing.
\newblock \emph{arXiv preprint arXiv:2311.16043}, 2023.

\bibitem[Gerig et~al.(2018)Gerig, Morel-Forster, Blumer, Egger, Luthi,
  Sch{\"o}nborn, and Vetter]{gerig2018morphable}
Thomas Gerig, Andreas Morel-Forster, Clemens Blumer, Bernhard Egger, Marcel
  Luthi, Sandro Sch{\"o}nborn, and Thomas Vetter.
\newblock Morphable face models-an open framework.
\newblock In \emph{2018 13th IEEE international conference on automatic face \&
  gesture recognition (FG 2018)}, pages 75--82. IEEE, 2018.

\bibitem[Ghosh et~al.(2011)Ghosh, Fyffe, Tunwattanapong, Busch, Yu, and
  Debevec]{ghosh2011multiview}
Abhijeet Ghosh, Graham Fyffe, Borom Tunwattanapong, Jay Busch, Xueming Yu, and
  Paul Debevec.
\newblock Multiview face capture using polarized spherical gradient
  illumination.
\newblock In \emph{Proceedings of the 2011 SIGGRAPH Asia Conference}, pages
  1--10, 2011.

\bibitem[Gotardo et~al.(2018)Gotardo, Riviere, Bradley, Ghosh, and
  Beeler]{gotardo2018practical}
Paulo Gotardo, J{\'e}r{\'e}my Riviere, Derek Bradley, Abhijeet Ghosh, and Thabo
  Beeler.
\newblock Practical dynamic facial appearance modeling and acquisition.
\newblock \emph{ACM Transactions on Graphics (ToG)}, 37\penalty0 (6):\penalty0
  1--13, 2018.

\bibitem[Han et~al.(2023)Han, Wang, and Xu]{han2023learning}
Yuxuan Han, Zhibo Wang, and Feng Xu.
\newblock Learning a 3d morphable face reflectance model from low-cost data.
\newblock In \emph{Proceedings of the IEEE/CVF Conference on Computer Vision
  and Pattern Recognition}, pages 8598--8608, 2023.

\bibitem[Han et~al.(2024)Han, Lyu, and Xu]{han2024high}
Yuxuan Han, Junfeng Lyu, and Feng Xu.
\newblock High-quality facial geometry and appearance capture at home.
\newblock In \emph{Proceedings of the IEEE/CVF Conference on Computer Vision
  and Pattern Recognition}, pages 697--707, 2024.

\bibitem[Hasselgren et~al.(2022)Hasselgren, Hofmann, and
  Munkberg]{hasselgren2022nvdiffrecmc}
Jon Hasselgren, Nikolai Hofmann, and Jacob Munkberg.
\newblock {Shape, Light, and Material Decomposition from Images using Monte
  Carlo Rendering and Denoising}.
\newblock \emph{arXiv:2206.03380}, 2022.

\bibitem[Jakob et~al.(2022)Jakob, Speierer, Roussel, Nimier-David, Vicini,
  Zeltner, Nicolet, Crespo, Leroy, and Zhang]{Mitsuba3}
Wenzel Jakob, Sébastien Speierer, Nicolas Roussel, Merlin Nimier-David, Delio
  Vicini, Tizian Zeltner, Baptiste Nicolet, Miguel Crespo, Vincent Leroy, and
  Ziyi Zhang.
\newblock Mitsuba 3 renderer, 2022.
\newblock https://mitsuba-renderer.org.

\bibitem[Kajiya(1986)]{kajiya1986rendering}
James~T Kajiya.
\newblock The rendering equation.
\newblock In \emph{Proceedings of the 13th annual conference on Computer
  graphics and interactive techniques}, pages 143--150, 1986.

\bibitem[Karis and Games(2013)]{karis2013real}
Brian Karis and Epic Games.
\newblock Real shading in unreal engine 4.
\newblock \emph{Proc. Physically Based Shading Theory Practice}, 4\penalty0
  (3):\penalty0 1, 2013.

\bibitem[Kautz et~al.(2000)Kautz, V{\'a}zquez, Heidrich, and
  Seidel]{kautz2000unified}
Jan Kautz, Pere-Pau V{\'a}zquez, Wolfgang Heidrich, and Hans-Peter Seidel.
\newblock A unified approach to prefiltered environment maps.
\newblock In \emph{Rendering Techniques 2000: Proceedings of the Eurographics
  Workshop in Brno, Czech Republic, June 26--28, 2000 11}, pages 185--196.
  Springer, 2000.

\bibitem[Ke et~al.(2022)Ke, Sun, Li, Yan, and Lau]{ke2022modnet}
Zhanghan Ke, Jiayu Sun, Kaican Li, Qiong Yan, and Rynson~WH Lau.
\newblock Modnet: Real-time trimap-free portrait matting via objective
  decomposition.
\newblock In \emph{Proceedings of the AAAI Conference on Artificial
  Intelligence}, pages 1140--1147, 2022.

\bibitem[Kelemen and Szirmay-Kalos(2001)]{kelemen2001microfacet}
Csaba Kelemen and Laszlo Szirmay-Kalos.
\newblock A microfacet based coupled specular-matte brdf model with importance
  sampling.
\newblock In \emph{Eurographics (short presentations)}, 2001.

\bibitem[Kerbl et~al.(2023)Kerbl, Kopanas, Leimk{\"u}hler, and
  Drettakis]{kerbl3Dgaussians}
Bernhard Kerbl, Georgios Kopanas, Thomas Leimk{\"u}hler, and George Drettakis.
\newblock 3d gaussian splatting for real-time radiance field rendering.
\newblock \emph{ACM Transactions on Graphics}, 42\penalty0 (4), 2023.

\bibitem[Laine et~al.(2020)Laine, Hellsten, Karras, Seol, Lehtinen, and
  Aila]{Laine2020diffrast}
Samuli Laine, Janne Hellsten, Tero Karras, Yeongho Seol, Jaakko Lehtinen, and
  Timo Aila.
\newblock Modular primitives for high-performance differentiable rendering.
\newblock \emph{ACM Transactions on Graphics}, 39\penalty0 (6), 2020.

\bibitem[Lattas et~al.(2020)Lattas, Moschoglou, Gecer, Ploumpis, Triantafyllou,
  Ghosh, and Zafeiriou]{lattas2020avatarme}
Alexandros Lattas, Stylianos Moschoglou, Baris Gecer, Stylianos Ploumpis,
  Vasileios Triantafyllou, Abhijeet Ghosh, and Stefanos Zafeiriou.
\newblock Avatarme: Realistically renderable 3d facial reconstruction"
  in-the-wild".
\newblock In \emph{Proceedings of the IEEE/CVF conference on computer vision
  and pattern recognition}, pages 760--769, 2020.

\bibitem[Lattas et~al.(2022)Lattas, Lin, Kannan, Ozturk, Filipi, Guarnera,
  Chawla, and Ghosh]{lattas2022practical}
Alexandros Lattas, Yiming Lin, Jayanth Kannan, Ekin Ozturk, Luca Filipi,
  Giuseppe~Claudio Guarnera, Gaurav Chawla, and Abhijeet Ghosh.
\newblock Practical and scalable desktop-based high-quality facial capture.
\newblock In \emph{European Conference on Computer Vision}, pages 522--537.
  Springer, 2022.

\bibitem[Li et~al.(2018)Li, Aittala, Durand, and Lehtinen]{Li:2018:DMC}
Tzu-Mao Li, Miika Aittala, Fr{\'e}do Durand, and Jaakko Lehtinen.
\newblock Differentiable monte carlo ray tracing through edge sampling.
\newblock \emph{ACM Trans. Graph. (Proc. SIGGRAPH Asia)}, 37\penalty0
  (6):\penalty0 222:1--222:11, 2018.

\bibitem[Liang et~al.(2024{\natexlab{a}})Liang, Gojcic, Nimier-David, Acuna,
  Vijaykumar, Fidler, and Wang]{liang2024photorealistic}
Ruofan Liang, Zan Gojcic, Merlin Nimier-David, David Acuna, Nandita Vijaykumar,
  Sanja Fidler, and Zian Wang.
\newblock Photorealistic object insertion with diffusion-guided inverse
  rendering.
\newblock \emph{arXiv preprint}, 2024{\natexlab{a}}.

\bibitem[Liang et~al.(2024{\natexlab{b}})Liang, Zhang, Feng, Shan, and
  Jia]{liang2024gs}
Zhihao Liang, Qi Zhang, Ying Feng, Ying Shan, and Kui Jia.
\newblock Gs-ir: 3d gaussian splatting for inverse rendering.
\newblock In \emph{Proceedings of the IEEE/CVF Conference on Computer Vision
  and Pattern Recognition}, pages 21644--21653, 2024{\natexlab{b}}.

\bibitem[Lyu et~al.(2023)Lyu, Tewari, Habermann, Saito, Zollh{\"o}fer,
  Leimk{\"u}ehler, and Theobalt]{lyu2023dpi}
Linjie Lyu, Ayush Tewari, Marc Habermann, Shunsuke Saito, Michael
  Zollh{\"o}fer, Thomas Leimk{\"u}ehler, and Christian Theobalt.
\newblock Diffusion posterior illumination for ambiguity-aware inverse
  rendering.
\newblock \emph{ACM Transactions on Graphics}, 42\penalty0 (6), 2023.

\bibitem[MacRobert(1928)]{wo1928spherical}
T.~M. MacRobert.
\newblock Spherical harmonics: An elementary treatise on harmonic functions and
  their applications, 1928.

\bibitem[McAllister et~al.(2002)McAllister, Lastra, and
  Heidrich]{mcallister2002efficient}
David~K McAllister, Anselmo Lastra, and Wolfgang Heidrich.
\newblock Efficient rendering of spatial bi-directional reflectance
  distribution functions.
\newblock In \emph{Proceedings of the ACM SIGGRAPH/EUROGRAPHICS conference on
  Graphics hardware}, pages 79--88, 2002.

\bibitem[Mildenhall et~al.(2021)Mildenhall, Srinivasan, Tancik, Barron,
  Ramamoorthi, and Ng]{mildenhall2021nerf}
Ben Mildenhall, Pratul~P Srinivasan, Matthew Tancik, Jonathan~T Barron, Ravi
  Ramamoorthi, and Ren Ng.
\newblock Nerf: Representing scenes as neural radiance fields for view
  synthesis.
\newblock \emph{Communications of the ACM}, 65\penalty0 (1):\penalty0 99--106,
  2021.

\bibitem[M{\"u}ller et~al.(2022)M{\"u}ller, Evans, Schied, and
  Keller]{muller2022instant}
Thomas M{\"u}ller, Alex Evans, Christoph Schied, and Alexander Keller.
\newblock Instant neural graphics primitives with a multiresolution hash
  encoding.
\newblock \emph{ACM transactions on graphics (TOG)}, 41\penalty0 (4):\penalty0
  1--15, 2022.

\bibitem[Munkberg et~al.(2022)Munkberg, Hasselgren, Shen, Gao, Chen, Evans,
  M\"uller, and Fidler]{Munkberg_2022_CVPR}
Jacob Munkberg, Jon Hasselgren, Tianchang Shen, Jun Gao, Wenzheng Chen, Alex
  Evans, Thomas M\"uller, and Sanja Fidler.
\newblock {Extracting Triangular 3D Models, Materials, and Lighting From
  Images}.
\newblock In \emph{Proceedings of the IEEE/CVF Conference on Computer Vision
  and Pattern Recognition (CVPR)}, pages 8280--8290, 2022.

\bibitem[Nguyen(2007)]{gpugems3}
Hubert Nguyen.
\newblock \emph{GPU Gems 3}.
\newblock Addison-Wesley Professional, 2007.

\bibitem[Nicolet et~al.(2021)Nicolet, Jacobson, and Jakob]{nicolet2021large}
Baptiste Nicolet, Alec Jacobson, and Wenzel Jakob.
\newblock Large steps in inverse rendering of geometry.
\newblock \emph{ACM Transactions on Graphics (TOG)}, 40\penalty0 (6):\penalty0
  1--13, 2021.

\bibitem[Papantoniou et~al.(2023)Papantoniou, Lattas, Moschoglou, and
  Zafeiriou]{papantoniou2023relightify}
Foivos~Paraperas Papantoniou, Alexandros Lattas, Stylianos Moschoglou, and
  Stefanos Zafeiriou.
\newblock Relightify: Relightable 3d faces from a single image via diffusion
  models.
\newblock In \emph{Proceedings of the IEEE/CVF International Conference on
  Computer Vision}, pages 8806--8817, 2023.

\bibitem[Parker et~al.(2010)Parker, Bigler, Dietrich, Friedrich, Hoberock,
  Luebke, McAllister, McGuire, Morley, Robison, et~al.]{parker2010optix}
Steven~G Parker, James Bigler, Andreas Dietrich, Heiko Friedrich, Jared
  Hoberock, David Luebke, David McAllister, Morgan McGuire, Keith Morley,
  Austin Robison, et~al.
\newblock Optix: a general purpose ray tracing engine.
\newblock \emph{Acm transactions on graphics (tog)}, 29\penalty0 (4):\penalty0
  1--13, 2010.

\bibitem[Qian(2024)]{qian2024versatile}
Shenhan Qian.
\newblock Versatile head alignment with adaptive appearance priors.
\newblock 2024.

\bibitem[Rainer et~al.(2023)Rainer, Bridgeman, and Ghosh]{rainer2023neural}
Gilles Rainer, Lewis Bridgeman, and Abhijeet Ghosh.
\newblock Neural shading fields for efficient facial inverse rendering.
\newblock In \emph{Computer Graphics Forum}, page e14943. Wiley Online Library,
  2023.

\bibitem[Riviere et~al.(2020)Riviere, Gotardo, Bradley, Ghosh, and
  Beeler]{riviere2020single}
J{\'e}r{\'e}my Riviere, Paulo~FU Gotardo, Derek Bradley, Abhijeet Ghosh, and
  Thabo Beeler.
\newblock Single-shot high-quality facial geometry and skin appearance capture.
\newblock \emph{ACM Trans. Graph.}, 39\penalty0 (4):\penalty0 81, 2020.

\bibitem[Saito et~al.(2024)Saito, Schwartz, Simon, Li, and
  Nam]{saito2024relightable}
Shunsuke Saito, Gabriel Schwartz, Tomas Simon, Junxuan Li, and Giljoo Nam.
\newblock Relightable gaussian codec avatars.
\newblock In \emph{Proceedings of the IEEE/CVF Conference on Computer Vision
  and Pattern Recognition}, pages 130--141, 2024.

\bibitem[Sarkar et~al.(2023)Sarkar, B{\"u}hler, Li, Wang, Vicini, Riviere,
  Zhang, Orts-Escolano, Gotardo, Beeler, et~al.]{sarkar2023litnerf}
Kripasindhu Sarkar, Marcel~C B{\"u}hler, Gengyan Li, Daoye Wang, Delio Vicini,
  J{\'e}r{\'e}my Riviere, Yinda Zhang, Sergio Orts-Escolano, Paulo Gotardo,
  Thabo Beeler, et~al.
\newblock Litnerf: Intrinsic radiance decomposition for high-quality view
  synthesis and relighting of faces.
\newblock In \emph{SIGGRAPH Asia 2023 Conference Papers}, pages 1--11, 2023.

\bibitem[Smith et~al.(2020)Smith, Seck, Dee, Tiddeman, Tenenbaum, and
  Egger]{smith2020morphable}
William A.~P. Smith, Alassane Seck, Hannah Dee, Bernard Tiddeman, Joshua
  Tenenbaum, and Bernhard Egger.
\newblock A morphable face albedo model.
\newblock In \emph{Proc. of the IEEE Conference on Computer Vision and Pattern
  Recognition (CVPR)}, pages 5011--5020, 2020.

\bibitem[Srinivasan et~al.(2021)Srinivasan, Deng, Zhang, Tancik, Mildenhall,
  and Barron]{srinivasan2021nerv}
Pratul~P Srinivasan, Boyang Deng, Xiuming Zhang, Matthew Tancik, Ben
  Mildenhall, and Jonathan~T Barron.
\newblock Nerv: Neural reflectance and visibility fields for relighting and
  view synthesis.
\newblock In \emph{Proceedings of the IEEE/CVF Conference on Computer Vision
  and Pattern Recognition}, pages 7495--7504, 2021.

\bibitem[Veach and Guibas(1995)]{veach1995optimally}
Eric Veach and Leonidas~J Guibas.
\newblock Optimally combining sampling techniques for monte carlo rendering.
\newblock In \emph{Proceedings of the 22nd annual conference on Computer
  graphics and interactive techniques}, pages 419--428, 1995.

\bibitem[Wang et~al.(2021)Wang, Liu, Liu, Theobalt, Komura, and
  Wang]{wang2021neus}
Peng Wang, Lingjie Liu, Yuan Liu, Christian Theobalt, Taku Komura, and Wenping
  Wang.
\newblock Neus: Learning neural implicit surfaces by volume rendering for
  multi-view reconstruction.
\newblock \emph{arXiv preprint arXiv:2106.10689}, 2021.

\bibitem[Wang et~al.(2023)Wang, Holynski, Zhang, and Zhang]{wang2023sunstage}
Yifan Wang, Aleksander Holynski, Xiuming Zhang, and Xuaner Zhang.
\newblock Sunstage: Portrait reconstruction and relighting using the sun as a
  light stage.
\newblock In \emph{Proceedings of the IEEE/CVF Conference on Computer Vision
  and Pattern Recognition}, pages 20792--20802, 2023.

\bibitem[Wang et~al.(2004)Wang, Bovik, Sheikh, and Simoncelli]{wang2004image}
Zhou Wang, Alan~C Bovik, Hamid~R Sheikh, and Eero~P Simoncelli.
\newblock Image quality assessment: from error visibility to structural
  similarity.
\newblock \emph{IEEE transactions on image processing}, 13\penalty0
  (4):\penalty0 600--612, 2004.

\bibitem[Wu et~al.(2024)Wu, Sun, Lai, Ma, Kobbelt, and Gao]{wu2024deferredgs}
Tong Wu, Jia-Mu Sun, Yu-Kun Lai, Yuewen Ma, Leif Kobbelt, and Lin Gao.
\newblock Deferredgs: Decoupled and editable gaussian splatting with deferred
  shading.
\newblock \emph{arXiv preprint arXiv:2404.09412}, 2024.

\bibitem[Xi et~al.(2024)Xi, Sida, Dongchen, Yuan, Bowen, Chengfei, and
  Xiaowei]{chen2024intrinsicanything}
Chen Xi, Peng Sida, Yang Dongchen, Liu Yuan, Pan Bowen, Lv Chengfei, and Zhou.
  Xiaowei.
\newblock Intrinsicanything: Learning diffusion priors for inverse rendering
  under unknown illumination.
\newblock \emph{arxiv: 2404.11593}, 2024.

\bibitem[Xu et~al.(2023)Xu, Zoss, Chandran, Gross, Bradley, and
  Gotardo]{Xu_2023_ICCV}
Yingyan Xu, Gaspard Zoss, Prashanth Chandran, Markus Gross, Derek Bradley, and
  Paulo Gotardo.
\newblock Renerf: Relightable neural radiance fields with nearfield lighting.
\newblock In \emph{Proceedings of the IEEE/CVF International Conference on
  Computer Vision (ICCV)}, pages 22581--22591, 2023.

\bibitem[Xu et~al.(2024)Xu, Chandran, Weiss, Gross, Zoss, and
  Bradley]{Xu_2024_CVPR}
Yingyan Xu, Prashanth Chandran, Sebastian Weiss, Markus Gross, Gaspard Zoss,
  and Derek Bradley.
\newblock Artist-friendly relightable and animatable neural heads.
\newblock In \emph{Proceedings of the IEEE/CVF Conference on Computer Vision
  and Pattern Recognition (CVPR)}, pages 2457--2467, 2024.

\bibitem[Yan et~al.(2012)Yan, Zhou, Xu, and Wang]{Yan2012-cq}
Ling-Qi Yan, Yahan Zhou, Kun Xu, and Rui Wang.
\newblock Accurate translucent material rendering under spherical gaussian
  lights.
\newblock \emph{Comput. Graph. Forum}, 31\penalty0 (7):\penalty0 2267--2276,
  2012.

\bibitem[Yang et~al.(2023)Yang, Zheng, Feng, Huang, Lai, Wan, Wang, and
  Ma]{yang2023towards}
Haotian Yang, Mingwu Zheng, Wanquan Feng, Haibin Huang, Yu-Kun Lai, Pengfei
  Wan, Zhongyuan Wang, and Chongyang Ma.
\newblock Towards practical capture of high-fidelity relightable avatars.
\newblock In \emph{SIGGRAPH Asia 2023 Conference Proceedings}, 2023.

\bibitem[Zhang et~al.(2021{\natexlab{a}})Zhang, Luan, Wang, Bala, and
  Snavely]{physg2021}
Kai Zhang, Fujun Luan, Qianqian Wang, Kavita Bala, and Noah Snavely.
\newblock {PhySG}: {I}nverse rendering with spherical gaussians for
  physics-based material editing and relighting.
\newblock In \emph{The IEEE/CVF Conference on Computer Vision and Pattern
  Recognition (CVPR)}, 2021{\natexlab{a}}.

\bibitem[Zhang et~al.(2018)Zhang, Isola, Efros, Shechtman, and
  Wang]{zhang2018unreasonable}
Richard Zhang, Phillip Isola, Alexei~A Efros, Eli Shechtman, and Oliver Wang.
\newblock The unreasonable effectiveness of deep features as a perceptual
  metric.
\newblock In \emph{Proceedings of the IEEE conference on computer vision and
  pattern recognition}, pages 586--595, 2018.

\bibitem[Zhang et~al.(2021{\natexlab{b}})Zhang, Srinivasan, Deng, Debevec,
  Freeman, and Barron]{zhang2021nerfactor}
Xiuming Zhang, Pratul~P Srinivasan, Boyang Deng, Paul Debevec, William~T
  Freeman, and Jonathan~T Barron.
\newblock Nerfactor: Neural factorization of shape and reflectance under an
  unknown illumination.
\newblock \emph{ACM Transactions on Graphics (ToG)}, 40\penalty0 (6):\penalty0
  1--18, 2021{\natexlab{b}}.

\bibitem[Zheng et~al.(2023)Zheng, Zhang, Yang, and Huang]{zheng2023neuface}
Mingwu Zheng, Haiyu Zhang, Hongyu Yang, and Di Huang.
\newblock Neuface: Realistic 3d neural face rendering from multi-view images.
\newblock In \emph{Proceedings of the IEEE/CVF Conference on Computer Vision
  and Pattern Recognition}, pages 16868--16877, 2023.

\bibitem[Zhu et~al.(2024)Zhu, Wang, and Yang]{zhu2024gs}
Zuo-Liang Zhu, Beibei Wang, and Jian Yang.
\newblock Gs-ror: 3d gaussian splatting for reflective object relighting via
  sdf priors.
\newblock \emph{arXiv e-prints}, pages arXiv--2406, 2024.

\end{thebibliography}
}
\clearpage
\setcounter{page}{1}
\maketitlesupplementary

In this supplementary document we begin by discussing the details of our capture protocol and initial tracking steps in \secref{sec:dataset}.  We provide more implementation details in \secref{sec:implAppendix}.  In \secref{sec:exp-detail} we highlight the particulars of our implementation of the related methods we use for comparisons, and elaborate on the synthetic dataset we used for evaluation.  Finally, we add additional results and failure cases in \secref{sec:more-results}.

\section{Dataset Details}
\label{sec:dataset}


\subsection{Capture Protocol}
\begin{figure}[!ht]
	\centering
	\includegraphics[trim=0 0 0 0, clip, width=\linewidth]{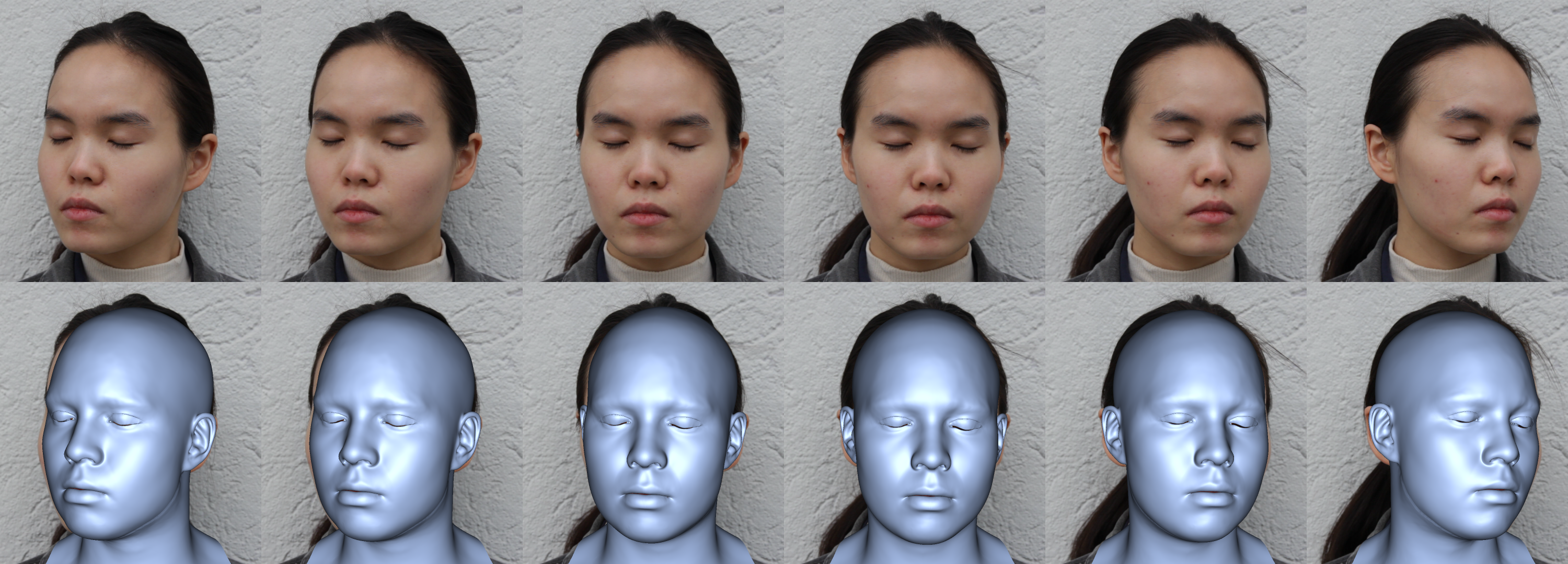}
	\caption{Example frames and initial tracking of a dataset. The first row shows in the input frames and the second row is the initial tracked geometry overlaid on the input.}
	\label{fig:data}
\end{figure}

We record videos at 25 fps using a Canon EOS 1200D camera fixed on a tripod. Depending on the distance to the subject, the camera is mounted with either a 35mm or 60mm lens and we use the corresponding focal length as a known parameter when calibrating the camera. The subjects are asked to slowly rotate their head 20 to 30 degrees to the left, right, up, and down directions. We do not record large head rotations as we noticed that the estimated head poses tend to be inaccurate on side view. \cref{fig:data} (row 1) shows some example frames of a dataset, where the left-most and right-most frames are the most extreme head rotations in this sequence. Note that even though we do not see a side face silhouette in the data, our method can recover the correct nose shape from shading as we have shown in \cref{fig:blendweights}. The datasets we captured span across multiple days at different locations.

\subsection{Initial Tracking} 
In the initial tracking stage, we estimate an initial mesh and per frame head poses. The initial mesh is parameterized using blendweights of a PCA face basis computed from the dataset of Chandran \etal \cite{chandran2020semantic}, which includes 50 eigen faces for identity and 25 for expression. Our tracking algorithm uses the combination of a landmark \cite{chandran2023continuous} loss  and a photometric loss, similar to Qian~\cite{qian2024versatile}. The only difference is that we solve for a global expression code since we assume the expression does not change in the same sequence. 
We apply a weight of 100 on the landmark loss and a weight of 30 on the photometric loss for all our datasets.

We further obtain an initial albedo estimate by averaging the projected texture across all the frames, and then we compute a piece-wise constant version based on a predefined face segmentation as in Rainer et al.~\cite{rainer2023neural}. The specular intensity map is initialized as a grayscale version of the initial diffuse albedo.

\section{Implementation Details}
\label{sec:implAppendix}
We capture 500 to 800 frames for each subject. We then uniformly sample around 250 frames to use in the inverse rendering. We did not observe a performance gain or drop when using all of the frames. The images are cropped to 1K resolution. We also solve the texture maps at 1K resolution. The environment map is a cubemap with resolution $6 \times 256 \times 256$ at the largest mip-level, and resolution $6 \times 8 \times 8$ at the smallest mip-level. For each pixel, we draw $256$ light samples and $256$ BRDF (cosine) samples for the diffuse render, and $64$ samples to estimate the view-dependent specular visibility. We employ the Adam optimizer~\cite{diederik2014adam} with a learning of $0.1$ for vertex positions and the environment map, and $0.001$ for the textures. We use the differentiable rasterizer from Laine~\etal~\cite{Laine2020diffrast} to obtain the primary visibility and the OptiX~\cite{parker2010optix} engine for ray tracing. Each subject is trained for $6000$ iterations which takes around 2 hours on a Nvidia RTX 3090 GPU. The weights for~\eqnref{eq:loss} are defined as
\begin{equation}
\begin{aligned}
& \lambda_{\text{mask}} = 0.1, \lambda_{\text{Lap}} = 10, \lambda_{\text{light}} = 0.1, \\
& \lambda_{\text{rough}} = 0.1, \lambda_{\text{diffuse}} = 0.01.
\end{aligned}
\end{equation}

\section{Experiment Details}
\label{sec:exp-detail}

\subsection{Implementation of the Related Methods}
Next we describe the steps performed to run the comparisons.
We use the default parameters of FLARE \cite{bharadwaj2023flare}. We noticed that the FLARE geometry is very bumpy, hence we tried setting a larger weight on the Laplacian mesh regularizer. However, this resulted in flatter face geometry and did not improve the results. 

For NextFace \cite{dib2021practical, dib2021towards, dib2022s2f2}, we obtained the best results in our experiments using only three frames that cover the whole face region. A similar behavior was observed by Azinovi\'c \etal \cite{azinovic2023high} in their experiments.

The original capture protocol used in SunStage \cite{wang2023sunstage} has a different format compared to ours; they record only a frontal view with the person rotating 360 degrees in place. We thus adapt the preprocessing code to the one from FLARE when running SunStage. We also do not solve for the focal length of the camera and set it as the ground truth value. In our experiments, the shape does not change much from the initial DECA \cite{feng2021learning} result, in both the coarse alignment stage and the photometric optimization stage of SunStage. 

NextFace and SunStage have no code for relighting in their release, so we did not compare relighting performance against these two baselines. The statistics in \tabref{tab:quantitative} are averaged over frames used in the optimization, \ie, all the frames for our method, FLARE, SunStage, and only three frames for NextFace.

\subsection{Synthetic Dataset}

\begin{figure}[b]
	\centering
	\includegraphics[trim=0 0 0 0, clip, width=\linewidth]{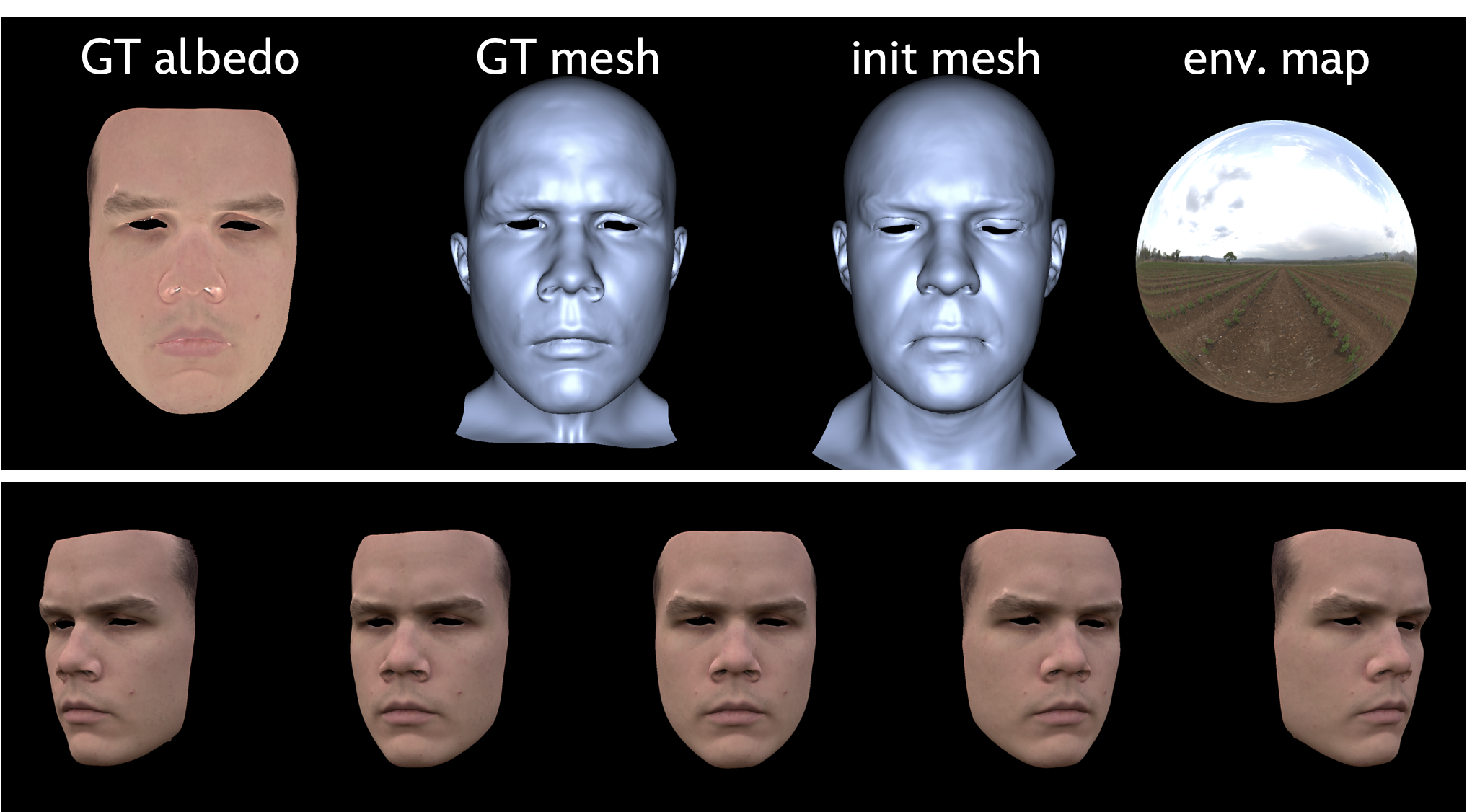}
	\caption{Assets and example frames of the synthetic dataset.}
	\label{fig:syn-data}
\end{figure}

We render a synthetic dataset with Lambertian material for the ablation study in \figref{fig:raytrace}. The assets, \ie ground truth diffuse albedo, mesh and environment maps are shown in row 1 of \figref{fig:syn-data}, and example frames are shown in row 2. The generated head poses are similar to those from a real dataset. We perform the same initial tracking algorithm using only the landmark loss on a front-facing frame. Instead of solving per frame head poses as for a real dataset, we use the ground truth head poses for the synthetic dataset in the inverse rendering stage. 

\section{Additional Results}
\label{sec:more-results}
Next we present additional results for the ablation and show some challenging situations for our algorithm.

\begin{figure}[tb]
	\centering
	\includegraphics[trim=0 0 0 0, clip, width=\linewidth]{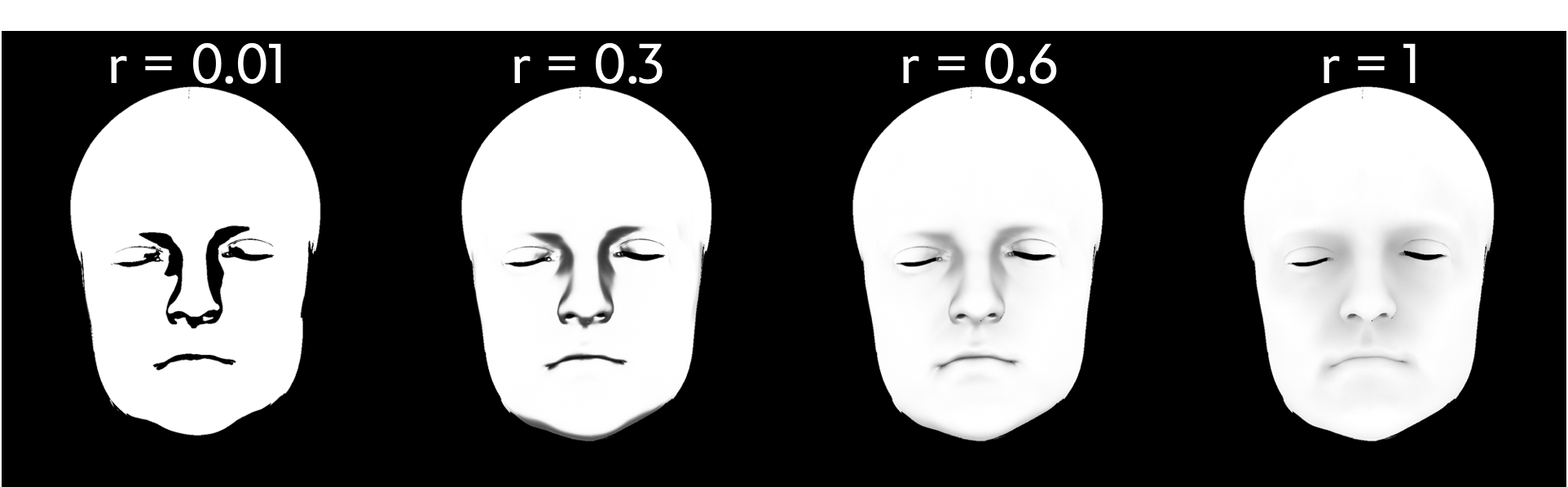}
	\caption{Visualization of the estimated view-dependent specular visibility term with different roughness values.}
	\label{fig:spec-vis}
\end{figure}

\begin{figure}[!ht]
	\centering
	\includegraphics[trim=0 0 0 0, clip, width=\linewidth]{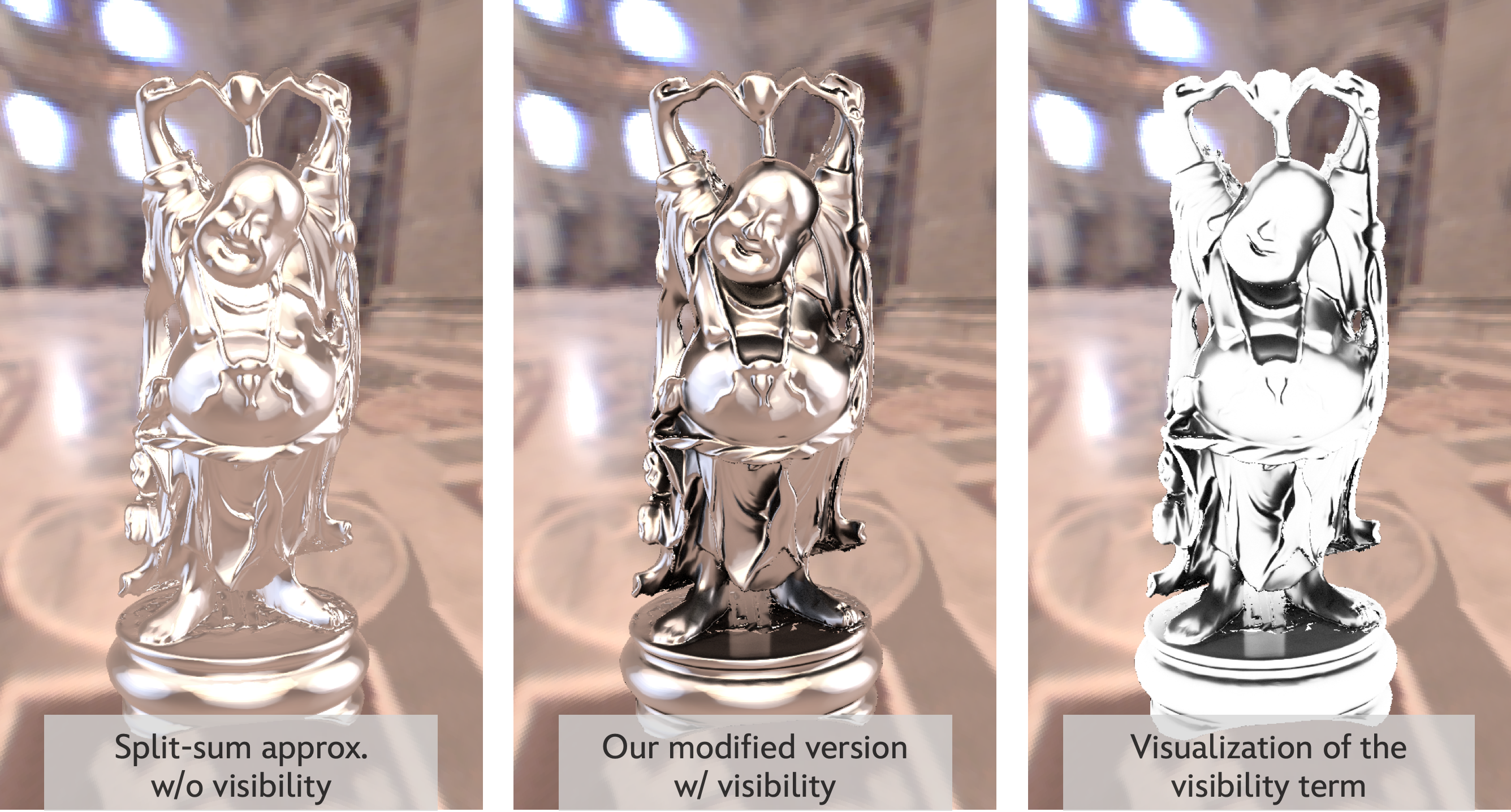}
	\caption{Rendering of a glossy Happy Buddha.}
	\label{fig:buddha}
\end{figure}

\paragraph{Visualization of the Visibility.}
First we provide additional visualizations of the view-dependent visibility under different roughness in \figref{fig:spec-vis}. This highlights the areas that are impacted by the visibility computation. When the roughness is small (mirror material), this visibility term is close to a binary mask and when the roughness is large (diffuse material), it gets closer to a view-independent ambient occlusion term. Note that the approximation error of ~\eqnref{eq:vis-approx} is small for a smaller roughness value and big for a larger roughness value. Applying the same approximation for the diffuse component would lead to a large error. We also show an example in~\figref{fig:buddha} of how our visibility-modulated split-sum approximation can be used in other rendering tasks as a practical way to add self-shadowing to glossy objects.

\begin{figure}[tb]
	\centering
	\includegraphics[trim=0 0 0 0, clip, width=\linewidth]{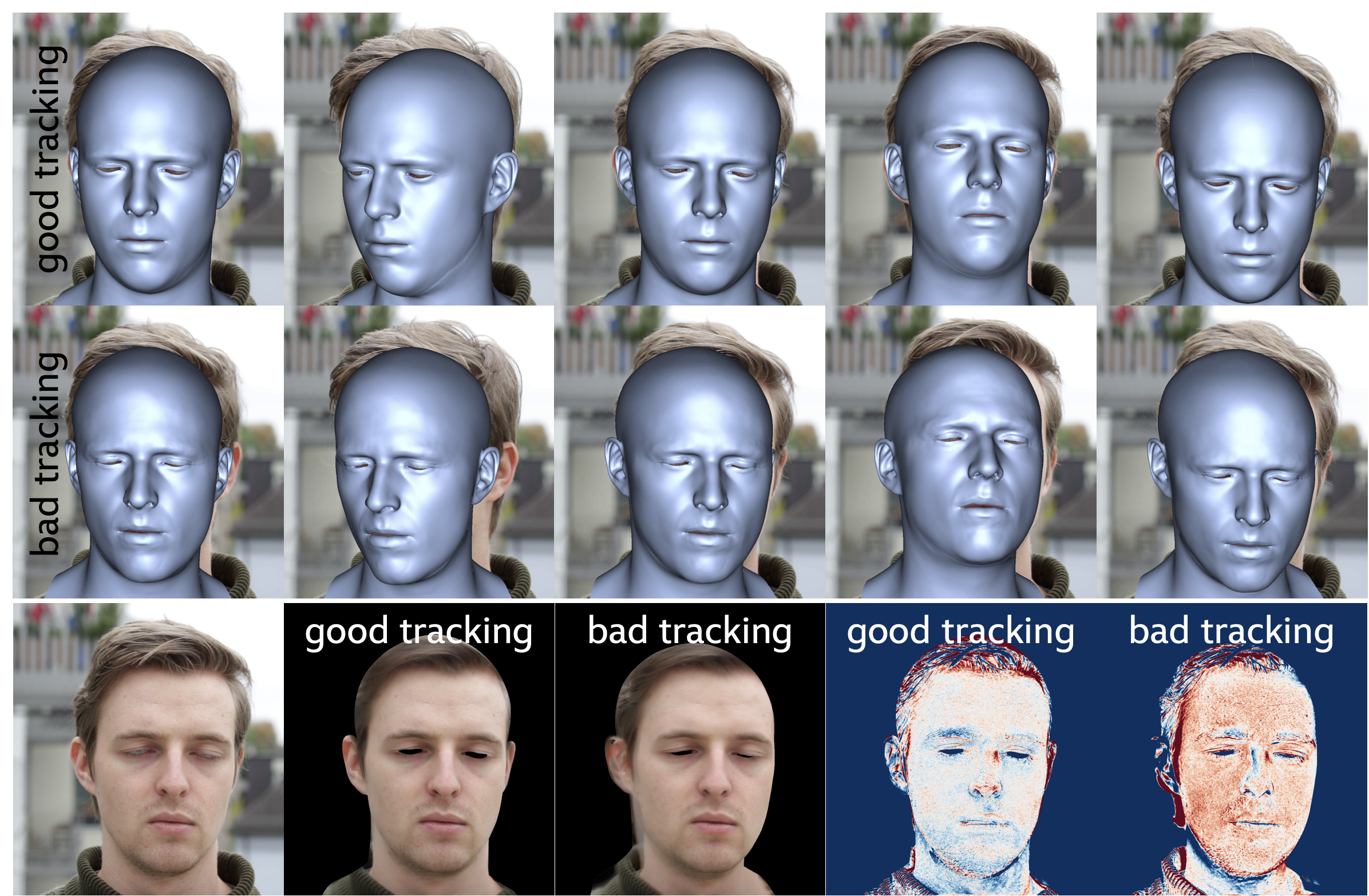}
	\caption{Results from good and bad head pose estimation from the initial tracking stage. The render error maps are displayed with a scale of -0.05~\includegraphics[width=2cm]{figures/divergence.png}~0.05.}
	\label{fig:fail-pose}
\end{figure}

\paragraph{Failure Cases.}
One of the limitations of our method is that we rely on good head pose estimation from the initial tracking stage.
If the initial tracking is incorrect or imprecise, see (\figref{fig:fail-pose} row 2), a misalignment of the render and the input image (\figref{fig:fail-pose} row 3 column 3) occurs, making the face appear distorted. 
We tried optimizing head poses in the inverse rendering stage but the results are often jittery over time and the textures are blurrier. Therefore, we decided to rely solely on the initial tracking for the head pose.
Note however, that our method still produces reasonable results even when the head poses are inaccurate. In these cases, our method explains the discrepancies between the tracked mesh and the input image using texture.

\begin{figure}[tb]
	\centering
	\includegraphics[trim=0 0 0 0, clip, width=\linewidth]{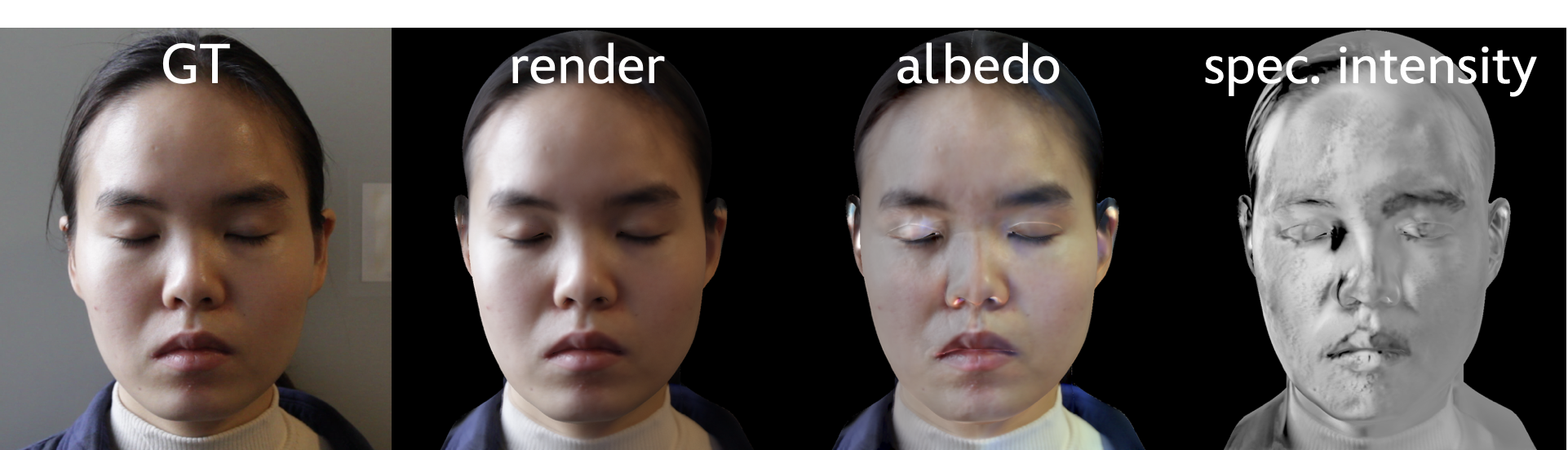}
	\caption{Poor appearance decomposition in challenging lighting conditions. While the render still matches the ground truth image, the albedo map contains some minor baked-in lighting.}
	\label{fig:fail-light}
\end{figure}

Although in theory our method can work under arbitrary static lighting conditions, there are challenging cases when our method still does not produce a good enough appearance decomposition. One such example is shown in \figref{fig:fail-light}. 
In this example, the left side of the face is overexposed while the right side is much darker in all frames. While the render still matches the input image, the reconstructed diffuse albedo and specular intensity maps contain a considerable amount of baked-in lighting. Note however, that our model still manages to disentangle a major part of the lighting from the appearance, \ie the brightness on the left and right sides of the diffuse albedo is similar. Capturing the same subject under multiple different lighting conditions can potentially improve the disentanglement \cite{baert2024spark}, which we leave as future work.

\begin{figure}[tb]
	\centering
	\includegraphics[trim=0 0 0 0, clip, width=\linewidth]{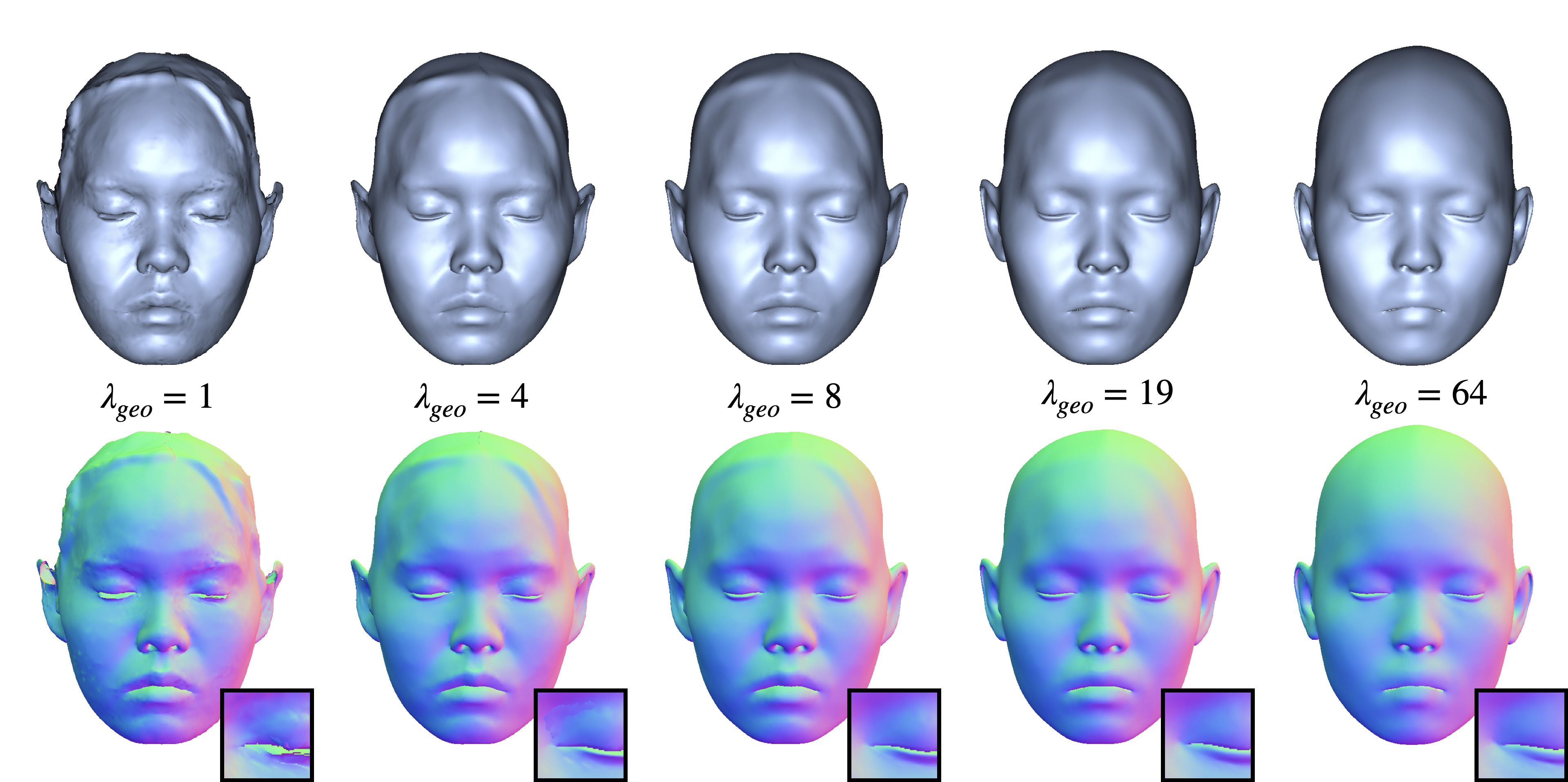}%
	\caption{Effects of $\lambda_{geo}$. The shaded meshes are shown on the top row with the corresponding normals visualized on the bottom.}
	\label{fig:lambda-geo}
\end{figure}

\paragraph{Effects of $\lambda_{geo}$.}
We use $\lambda_{geo} = 19$ based on~\cite{nicolet2021large}, but this smoothing term can be tuned for different subjects to get better geometry. An ablation for different values of $\lambda_{geo}$ is shown in \figref{fig:lambda-geo}. Note that a large $\lambda_{geo}$ leads to loss of details, but a small $\lambda_{geo}$ can lead to self-intersections.

\begin{figure}[tb]
	\centering
	\includegraphics[trim=0 0 0 0, clip, width=\linewidth]{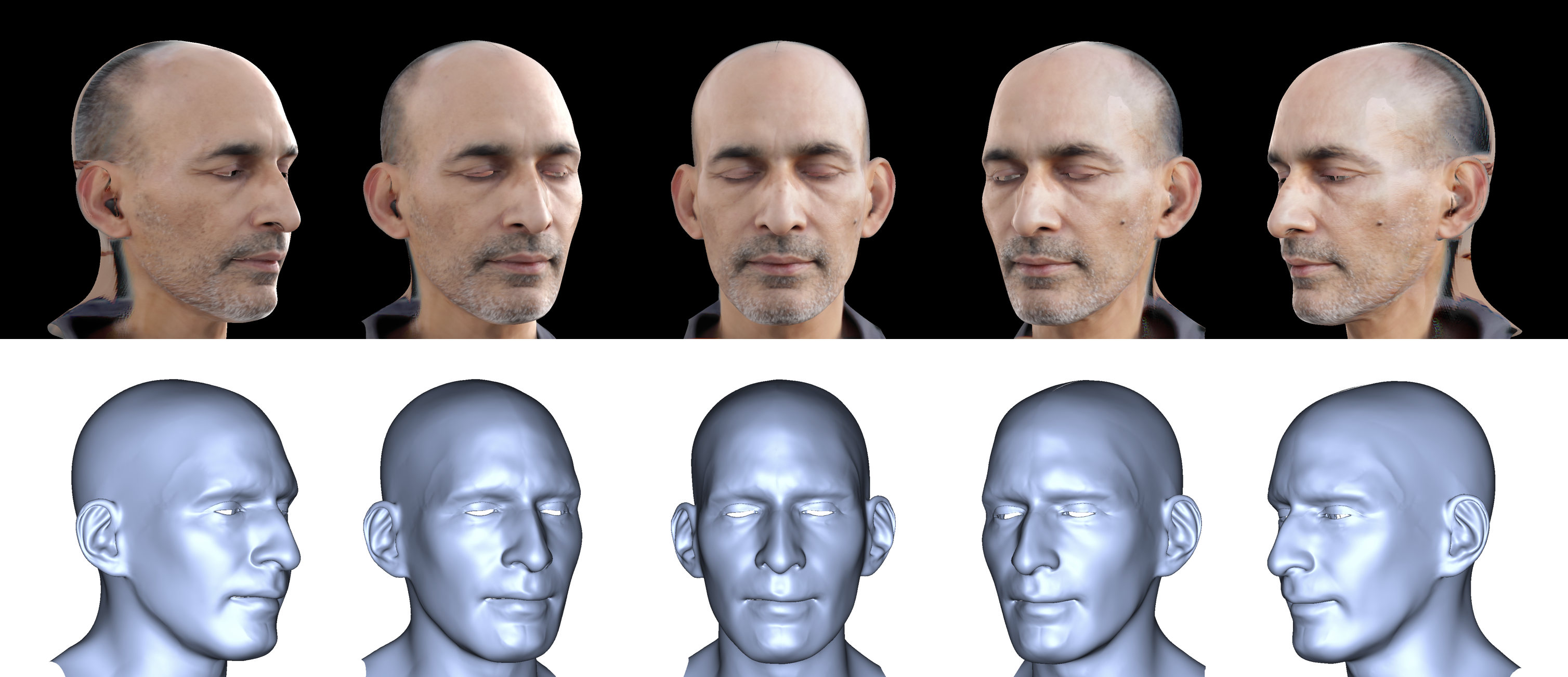}
	\caption{Novel views rendered in the capture environment (top row) along with the corresponding meshes (bottom row).}
	\label{fig:novel-views}
\end{figure}

\paragraph{Novel Views in the Capture Environment.}
We show novel view renders in the capture environment in~\figref{fig:novel-views}. We primarily focused on the facial regions for this project. Artifacts around the boundaries can be improved with some engineering efforts, such as better masking and better initial alignment. Reconstruction of the hair and shoulders is also an interesting area for future work, which can improve the overall quality.

\begin{figure}[!ht]
	\centering
	\includegraphics[trim=0 0 0 0, clip, width=\linewidth]{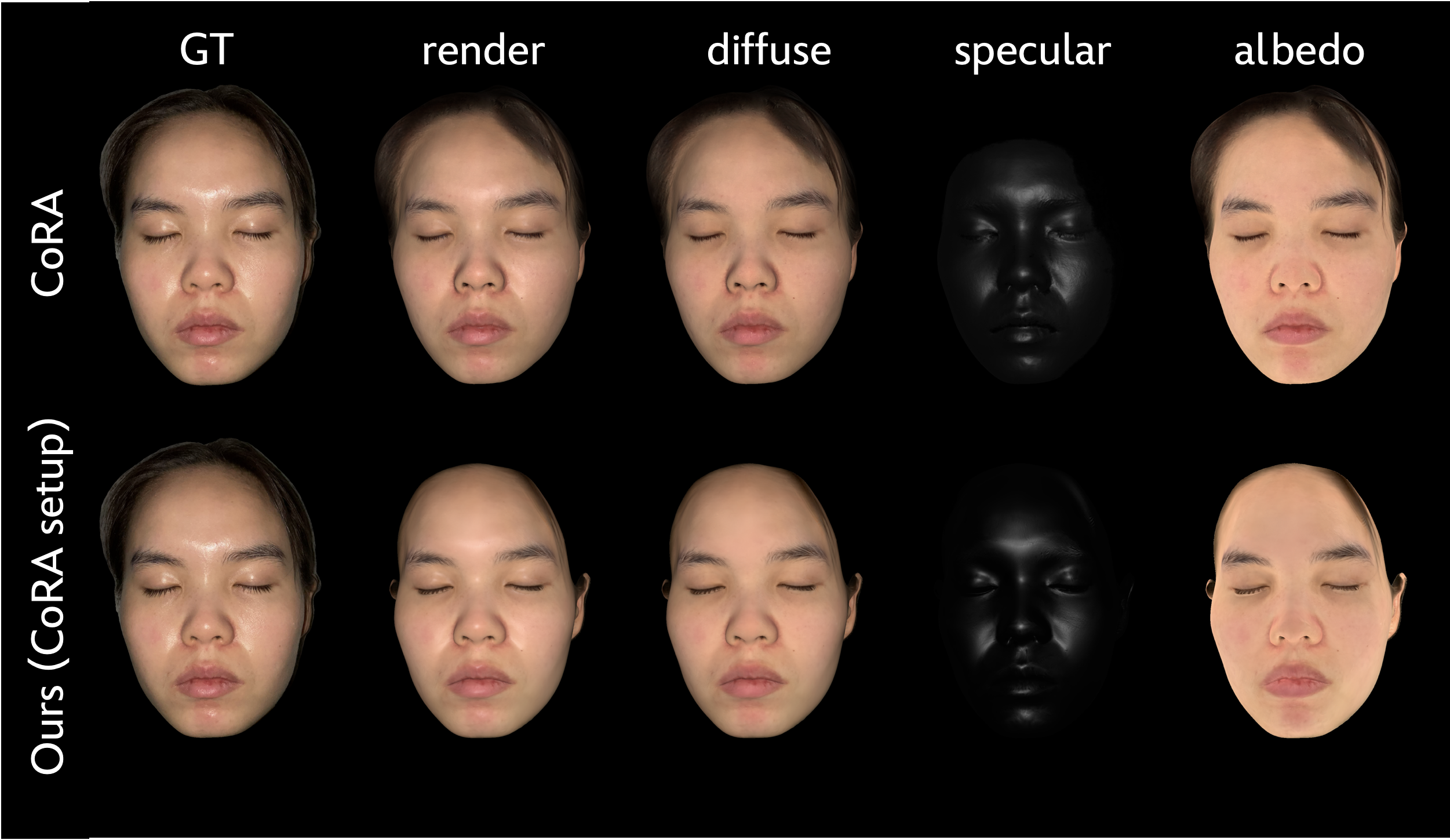}
	\caption{Comparison with CoRA on their data protocol.}
	\label{fig:cora}
\end{figure}

\paragraph{Comparison with CoRA.}
CoRA~\cite{han2024high} requires a more constrained capture setup with a co-located light and camera in a dark room. We ran our method and CoRA on a subject using the CoRA capture protocol, see~\figref{fig:cora}, demonstrating that our method achieves equally good reconstruction given the same data. Additionally, our method also works in more generic environments.

\end{document}